\documentclass[nonacm,authorversion,natbib=false,screen]{jair}  

\usepackage{url}
\usepackage{graphicx}
\usepackage{scalerel}
\usepackage{tabularx,booktabs}
\usepackage{multirow}
\usepackage{enumitem}
\usepackage{amsmath}
\usepackage{booktabs}
\usepackage{siunitx}
\def\smiley{\scalerel*{\includegraphics{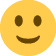}}{\textrm{\textbigcircle}}}
\DeclareMathOperator*{\argmax}{argmax}

\newcommand{\corpus}{\textsc{EmoWOZ-CS}\xspace}

\setcopyright{cc}
\copyrightyear{2025}

\RequirePackage[
  datamodel=acmdatamodel,
  style=acmnumeric,
  backend=biber,
  giveninits=true
  ]{biblatex}

\begin{document}

\title[Leveraging the Wizard of Oz Technique to Model Emotions in Customer Service Interactions]{A Customer Journey in the Land of Oz: Leveraging the Wizard of Oz Technique to Model Emotions in Customer Service Interactions}

\author{Sofie Labat}
\authornote{Corresponding Author.}
\orcid{0000-0003-1675-8927}
\email{sofie.labat@ugent.be}
\affiliation{
  \institution{LT3, Language and Translation Technology Team, Ghent University}
  \city{Ghent}
  \country{Belgium}
}

\author{Thomas Demeester}
\orcid{0000-0002-9901-5768}
\email{thomas.demeester@ugent.be}
\affiliation{
  \institution{IDLab, Internet technology and Data science Lab, Ghent University--imec}
  \city{Ghent}
  \country{Belgium}
  }

\author{Véronique Hoste}
\orcid{0000-0002-0539-4630}
\email{veronique.hoste@ugent.be}
\affiliation{
  \institution{LT3, Language and Translation Technology Team, Ghent University}
  \city{Ghent}
  \country{Belgium}
}

\renewcommand{\shortauthors}{Labat, Demeester \& Hoste}

\begin{abstract}
{\bf Background:} 
Emotion-aware customer service needs in-domain conversational data, richer annotations, and forward-looking inference, but existing resources for emotion recognition in conversation are often out-of-domain, narrowly labeled, and focused on post-hoc detection.

{\bf Objectives:}
We run a controlled Wizard of Oz (WOZ) experiment to elicit in-domain interactions with targeted affective trajectories. We aim to: (1) Evaluate WOZ experiment with operator-steered valence trajectories as a corpus design for emotion research; (2) Quantify human performance and variation in emotion annotation, including divergences between self-reports and third-party judgments; (3) Benchmark detection and forward-looking emotion inference in real-time customer support.

{\bf Methods:}
We collected \corpus, a bilingual (Dutch–English) corpus of 2,148 written customer service dialogues from 179 participants on scenarios across commercial aviation, e-commerce, online travel agencies, and telecommunication. Human \textit{wizards} posed as autonomous agents and were instructed to steer conversations toward target end valence (positive/neutral/negative). Participant turns are annotated with discrete emotions and valence–arousal–dominance (VAD) scores; wizard turns are multilabeled with operator response strategies. We post-edited English translations, anonymized data, and included demographics and Big Five traits. We analyzed distributions, inter-annotator agreement, self-report vs. third-party alignment, strategy–emotion interactions, and temporal emotion progression; we also benchmarked fine-tuned encoders for emotion detection and used a zero-shot LLM alongside encoders for forward-looking inference.

{\bf Results:}
Neutral dominates participant messages; desire and gratitude are the most frequent non-neutral emotions. Agreement is moderate for multilabel emotions and valence, lower for arousal and especially dominance; self-reports diverge notably from third-party labels, with highest alignment for neutral, gratitude, and anger. Objective strategies (explanation, request information, help online/offline) often elicit neutrality or gratitude, whereas suboptimal strategies (miscomprehension, non-collaborative/inappropriate response, irony) increase anger, annoyance, disappointment, desire, and confusion. Some affective strategies (cheerfulness, gratitude) foster positive reciprocity, whereas others (apology, empathy) are less restorative and often leave desire, anger, or annoyance. Temporal dynamics show successful conversation-level steering toward prescribed targets, most distinctly for negative trajectories; positive and neutral targets yield similar final valence distributions. Benchmarks indicate the challenging nature of forward-looking emotion inference based on prior participant and operator turns.

{\bf Conclusions:}
\corpus~advances emotion modeling by conducting a controlled, task-specific WOZ experiment. The dataset incorporates both discrete and dimensional emotion labels, operator strategies, profile-specific metadata, and is accompanied by a predictive evaluation benchmark. Together, the corpus and accompanying analyses enable proactive, strategy-aware guidance for real-time customer support and provide a publicly available resource for research on emotion-aware dialogue systems. Our corpus and code\footnote{Code will be made publicly available upon manuscript acceptance.} are available at \url{https://lt3.ugent.be/resources/emowoz-cs-en-nl/}.
\end{abstract}


\maketitle

\section{Introduction}\label{sec1:intro}
Along with interaction goals, emotions are the hidden engines of human interaction, influencing the human decision-making process on what gets said, what gets heard, and what happens next~\cite{Lerner2015}. Service research shows that these affective undercurrents are not incidental: they are crucial in shaping customer satisfaction, loyalty, and downstream behaviors~\cite{Dobrev2022_big,DeCleen2025}. Evidence from utilitarian service settings further indicates that their impact is asymmetrical, with negative emotions depressing satisfaction and loyalty more than equivalent positive emotions increase them~\cite{Rychalski2017}. In customer support, accurately inferring and responding to emotion enhances communication quality and perceived service. While contact centers actively embrace generative artificial intelligence (AI) to transform customer care, humans remain essential for their ability to handle complex and emotionally nuanced interactions~\cite{Blackader2025}. Automatic emotion recognition in text remains a challenging task~\cite{Poria2019,Deng2023_survey,Pereira2024_deep}, especially in real-world interactions, where it can be used to provide, e.g., real-time de-escalation, personalization, and quality assurance. Nevertheless, recent studies in customer relationship management and conversational AI further underscore that detecting and adapting to consumer emotions is critical for improving agent--customer alignment and overall satisfaction, reinforcing the strategic value of integrating affective computing into customer support workflows~\cite{GracyTheresa2025}.

Against this backdrop, we identify three research gaps that prevent emotion-aware systems from delivering reliable, real-time value in customer support.
First, raw data for emotion recognition in conversation (ERC) is misaligned with the communicative realities of service dialogue. Popular benchmarks for ERC are drawn from second-language learner resources~\cite{Li2017} or scripted TV shows~\cite{Chen2018,Zahiri2018,Poria2019b} rather than unscripted, goal-driven exchanges in which institutional constraints, stakes, and task context shape how affect is expressed and perceived~\cite{Rafaeli1987}. This mismatch undermines ecological validity and inflates out-of-domain performance. In addition, high-quality customer interaction data are hard to obtain, as textual service logs are typically proprietary and thus unavailable as public training resources.

A second gap concerns the constrained annotation frameworks and limited meta-information in ERC corpora. Most ERC datasets lack self-reported emotion labels and omit interlocutor information such as demographics or personality measures. These omissions hinder perspectivist approaches~\cite{Frenda2025}, which examine how individual profiles and viewpoints drive variation in the emotional meaning assigned to salient stimuli. Such variation, evidenced by divergences between self-reports and third-party annotations~\cite{Labat2022a}, remains underexplored and continues to challenge text-based detection models~\cite{plaza-del-arco-etal-2024-emotion}. Moreover, service interactions frequently convey affect implicitly and exhibit naturally imbalanced label distributions dominated by neutral content~\cite{Labat2024}. 
To enable robust measurement, annotation frameworks should integrate discrete emotion categories with multidimensional affective scales (e.g., valence, arousal, dominance). At the same time, operator strategies should be captured in a distinct taxonomy that specifies communicative tactics for interpersonal emotion regulation, thereby influencing the trajectory of affect over the course of the interaction. 

For the final research gap, we address the challenging task of automatic emotion inference. Current approaches to ERC primarily focus on post-hoc labeling of emotions expressed in observed utterances~\cite{Labat-etal-2022-emotional,Zhang2025}, rather than forecasting the emotions likely to arise when the content of the next turn remains unknown~\cite{li-etal-2021-emo-inf}. We argue that in-text detection of expressed emotion is limited in its capacity to prevent escalation in real-time and offers little support for optimizing turn-by-turn decision-making in real-time support contexts. Such contexts, however, require forward-looking models that can anticipate how each conversational action will shape the customer's next emotional state and the overall trajectory of the interaction. Practical systems should therefore shift toward predictive emotion inference, enabling proactive guidance, timely intervention, and closer alignment with the temporal dynamics of service interactions.

To address these gaps, we introduce \corpus, a new corpus of customer service dialogues collected through a controlled Wizard of Oz (WOZ) experiment. In this setup, participants engaged in written conversations with what they believed to be an autonomous agent, while in reality each interaction was fully operated by a human \textit{wizard}. This design allowed us to capture in-domain, task-oriented dialogues while systematically steering the affective trajectory of the exchange. \corpus~makes several contributions: First, rather than passively relying on the initial valence of the conversation starter, wizards were explicitly instructed to actively steer the dialogue toward a predetermined affective outcome. The collected dialogues are annotated with a multi-layered emotion scheme that combines categorical labels with continuous dimensions of valence, arousal, and dominance, thereby supporting richer and more ecologically valid measurement than discrete categories alone. In addition, all wizard replies are tagged with operator strategy labels, capturing the communicative tactics employed to regulate customer emotions. To complement these annotations, participants were also asked to retrospectively label their first conversation with discrete emotion categories, providing self-reported insights into their own affective experience. Beyond utterance-level annotations, we also gathered participant profiling data, including demographic variables (age, gender), personality traits, familiarity with the WOZ setup, and prior experience with chatbots. Finally, unlike most prior Wizard of Oz corpora, which are restricted to English, \corpus~is available in both Dutch (Flemish) and English. Its bilingual nature enhances accessibility, fosters cross-linguistic comparison, and supports the development of multilingual emotion-aware dialogue systems. Taken together, these design choices make \corpus~a resource for both emotion recognition and forward-looking inference, supporting the study of how dialogue strategies anticipate, influence, and regulate customer emotions in real time. Building on these contributions, our study addresses three overarching research objectives: 
\begin{enumerate}[label=\textbf{(RO\arabic*)}]
    \item \textbf{Evaluate Wizard of Oz as a corpus design for emotion research.} We compare the WOZ paradigm with operator-steered valence trajectories to alternative collection methods and examine how operator strategies shape affective dynamics and what factors condition the presence of emotions over time.
    \item \textbf{Quantify human performance and variation in emotion annotation.} We assess agreement for dimensional (valence, arousal, dominance) and categorical labels, and analyze divergences between self-reports and third-party judgments to reveal perspectivist differences and establish realistic annotation baselines.
    \item \textbf{Benchmark machine learning performance on detection and forward-looking inference.} We compare fine-tuned encoders (e.g., RobBERT, ModernBERT) with a zero-shot LLM on discrete and dimensional emotion recognition, and test forward-looking inference to evaluate how well future emotional states can be anticipated in real-time customer service contexts.
\end{enumerate}

The remainder of this paper is structured as follows: Section~\ref{sec2:data} introduces related work on three principal paradigms to collect text corpora and their trade-offs. Section~\ref{sec3:data_collect} details the methodology and procedures of the \corpus~experiment and corpus creation, including scenarios, valence trajectories, and operator response strategies. Section~\ref{sec4:analysis} analyzes the corpus characteristics, participant profiles, emotion distributions, annotation agreement, and conversational dynamics (including emotion progressions). Section~\ref{sec5:ML} presents machine learning experiments on emotion inference, contrasting fine-tuned encoders with zero-shot large language models. Section~\ref{sec6:conclus} discusses our main findings with respect to the research objectives and concludes the paper by outlining its limitations and directions for future work.

\section{Three Methods to Collect Corpora}\label{sec2:data}
A machine learning system is only as reliable as the data it is trained on. Because our study presents a new corpus, it is essential to situate our methodology within broader data collection practices and justify our choice of the Wizard of Oz design. Training corpora can be obtained in three main ways: by harvesting real-world data, generating synthetic data with AI models, or recruiting human participants. While such approaches apply across modalities, we focus on text, as it is most relevant to our contribution. In the following, we outline the strengths and limitations of each method for collecting customer interactions and introduce the Wizard of Oz paradigm as an alternative strategy for eliciting dialogues with human subjects.

\subsection{Reusing Real-World Data}
Most corpora for machine learning are created by relying on existing data. The internet and the abundance of digitized text certainly contributed to the approach's popularity. Nevertheless, it is the rise of the Web 2.0 as a dynamic medium for human interaction that caused a corresponding rise in corpora of online user-generated content for tasks such as sentiment analysis~\cite{Pang-2004,Ganu-2009,Maas-2011,Pontiki2016}, user trait profiling~\cite{Stillwell2015,Plank-hovy-2015,Verhoeven-2016}, and argument mining~\cite{Boltuzic-2014,Park-2014,Habernal-2017}. The advantages of using user-generated content are clear: vast amounts of texts can be collected in a short period of time and the costs associated with the scraping process are low, even if in the case of supervised machine learning, some costs might be reserved for the annotation of data. 

The process also has disadvantages, one of them being the bias and inaccuracies contained in social data~\cite{Olteanu2019}. Mitigating bias, reducing stereotypes, and promoting fairness is an active area of research in AI~\cite{Mehrabi2021}, but even though the field is making progress~\cite{Zhao-2018,Yan2020}, the issues are far from solved. An example of this claim is the incident with Google's multimodal flagship model Gemini~\cite{Geminiteam2024} in February 2024, when it was critiqued for overcorrecting racial bias by rendering “inaccuracies in some historical image generation depictions”~\cite{Sriram2024}. Another issue is that not all data are freely available to scrape. The boom of generative AI brought about a disruptive shift in intellectual property rights, data privacy, and cybersecurity~\cite{Appel2023,Gupta2023}. 
In parallel, social media companies have introduced stricter policies to restrict scraping and safeguard user privacy, making platform-derived data increasingly uncertain for long-term usability.

Existing real-world data on customer interactions are mainly owned by large enterprises, while smaller businesses do not necessarily have the resources to build accurate in-house AI systems. One way to circumvent this lack is by turning to social network sites which contain message threads of customers requesting information, sharing positive experiences, or voicing complaints~\cite{Hadifar2021}. Prior research has shown that in the latter case, businesses try to redirect complaints away from the eyes of the public to private channels by, e.g., asking them to send a direct or private message~\cite{Einwiller2015,VanHerck2020}. In practice, scraping complaints from social media often yields shorter, unfinished conversations, as their endings are hidden in private chats~\cite{Labat2024}.

\subsection{Synthetic Corpus Generation Through AI}
A second and newer method to collect corpora is by generating the data with large language models (LLMs). Using generative AI to create data for machine learning might seem a bit of a detour, since generative models could also be used to build end-to-end applications that do, in principle, not require additional data. Nevertheless, numerous incidents have demonstrated that end-to-end LLM systems are at times hard to control in real-world applications: hallucinations from Air Canada's lying chatbot caused the company to lose a small claims court case against a passenger on bereavement fares~\cite{Garcia2024}; DPD's swearing chatbot wrote a haiku on how useless the company is~\cite{Clinton2024}. Corpora will thus continue to play an important role in the age of generative AI~\cite{Crosthwaite2023}.

Although language modeling has been around for a long time in the field of natural language processing (NLP)~\cite{Shannon1948,Bengio2003}, it is mostly due to the transformer architecture~\cite{Vaswani2017} that the idea of pretraining models to acquire general knowledge about language and the world from texts gained traction. While LLMs can be applied to any NLP task, their impact has been most transformative in tasks that involve the generation of text (e.g., text summarization~\cite{Wang2022}, code generation~\cite{Dehaerne2022}, chatbots~\cite{OpenAI2022}). OpenAI's family of GPT models~\cite{Brown2020,Openai2024gpt4} and their application ChatGPT, which brought these models to the general public, are linked to a broader generative AI explosion. Nowadays, many big tech companies have released their own generative LLMs (e.g., Google's Gemini~\cite{Chowdhery2023}, Facebook's Llama~\cite{Touvron2023llama}), while startups also jumped on the topic and released models such as Mistral's open-source Mixtral~\cite{jiang2024mixtral} or Anthropic's Claude~\cite{anthropic2024}. Even though earlier generative models operated on a single modality (e.g., text), newer models can often handle multimodal data (combinations of text, voice, images, and video) on the input and output level.

Similar to corpora of real-world data, datasets created by generative AI are collected in a short span of time and at low costs. While open-source models can be implemented for free, other LLMs might require a fee for their usage – the amount of which is determined by the number of input and output tokens. Data creation through generative AI is an accessible method in the sense that many LLMs can be consulted through web applications by writing prompts in natural language, although API keys are provided for developers. Finally, in contrast to the previous method, an LLM can be prompted to generate balanced datasets along a predefined set of labels. 
For the collection of customer interactions, this means that through proper prompting, an LLM could generate entire customer dialogues grounded on a diverse set of events.

While there are cost benefits to the use of generative AI, it also has certain drawbacks. Although LLMs perform well in lab settings, their reliability in specific real-world applications might not always hold up and it is up to their users to adequately distinguish between cases where these models are effective and cases where they are not~\cite{Brynjolfsson2023}. First, language models learn and replicate bias encountered in their training data~\cite{feng-etal-2023,kamruzzaman-etal-2024}. Moreover, generative AI systems are prone to hallucinations, producing false or misleading claims in highly convincing ways~\cite{Ji2023}. As for the creation of synthetic corpora through LLMs, it is important to postprocess automatically generated texts by flagging falsehoods, such that these instances are properly handled during training and their signals are not propagated at inference time. In this sense, flagged counterexamples (whether falsehoods or other types) are highly useful for explicitly training machine learning systems on what not to generate during inference~\cite{Dreossi2018}. For customer interactions, counterexamples most relevant to real-world applications are found in operator utterances: wrongly understanding and reacting to customer intents, being impolite or non-constructive, hallucinating fake company policies, etc.


Another shortcoming is the difference in distribution between synthetically generated content and real-world content: in the former, probable data points are overestimated, while improbable ones are underestimated. When LLMs are trained on AI-generated content, the models will collapse over time as the tails of their original distribution disappear, causing them to forget improbable data instances~\cite{Shumailov2023}. In practice, this might increase the chances of models hallucinating in their output when they come across underrepresented instances. Furthermore, some research has indicated that collaborative writing with the assistance of certain LLMs can lead to a decrease in lexical and content diversity~\cite{Padmakumar2024}. Similarly, LLM use among crowd workers on text summarization seems to result in high-quality but homogeneous responses~\cite{Veselovsky2023}. Other researchers have also acknowledged that maintaining diversity in synthetically generated data becomes increasingly challenging when scaling up~\cite{Cosmopedia2024}, although techniques to improve diversity in the models' output are also being explored~\cite{Hashimoto2019,Hayati2024}. Nevertheless, researchers remain cautious about using synthetically generated data~\cite{Kaddour2023}: while some model builders are enhancing real-world data with synthetic data, others have faced setbacks with pretraining models using synthetic data, as it resembled earlier models too closely~\cite{batch2024}. To conclude, synthetic data can augment existing customer interaction data, but the exploration of its potential for model training remains in its early stages. At this stage, more research and human effort in curating synthetic data are required.

\subsection{Studies With Human Participants and the Wizard of Oz Technique}

Many NLP tasks can be effectively performed using publicly available real-world content. However, tasks that focus on subjective, creative, or aspects of human behavior often benefit from human responses to questionnaires or participation in experiments. LLMs are not ideal for these tasks either, as they may lack sufficient human diversity and struggle with tasks involving narrow domains or social bias. Traditionally, studies involving human subjects were conducted in controlled lab settings. Nowadays, researchers in fields such as economics, cognitive sciences, social sciences, and AI increasingly use online recruitment services (e.g., Prolific, Amazon's MTurk) to find participants~\cite{Eerol2021}. These platforms not only expedite the data collection process, but also provide access to a diverse audience with varied demographics and profiling characteristics. Unlike the two previously mentioned collection methods, studies involving human participants typically result in smaller datasets and incur higher collection costs. Nonetheless, both online and offline studies offer researchers significant control over the corpus being collected and the settings in which the collection takes place. This also implies that considerable time and effort is to be spent on the design of the survey or experiment~\cite{Garcia-Molina2016}.

\subsubsection{Design of Studies Involving Human Subjects}
There are numerous studies that foster the creativity and commonsense knowledge of participants. Hossain et al.~\cite{hossain-etal-2019} instructed crowdsourced editors to make news headlines funny. The resulting dataset contains both the original headlines and their edits, meaning that both positive and negative instances of humor are available for research. Abu Farha et al.~\cite{Farha-etal-2022} gathered sarcastic texts in English and Arabic and requested the writers of these texts to generate counterexamples by rephrasing the original message in a non-sarcastic manner. They also asked writers to explain why the original text is sarcastic, which is information that can subsequently be used for explainability. A unique benefit of these types of studies is that participants who generate data can directly be asked to reflect on their own perspective, inner thoughts, and cognitive processes underlying the generation. In a study on detecting emotions and appraisals, Troiano et al.~\cite{Troiano2023} tasked participants to recollect an event that caused a particular emotion in them. The resulting event descriptions are annotated by both their writers and external readers, and analysis shows that readers have a higher agreement on appraisal values among themselves than with the actual writers of the events.

While the previously described surveys present innovative questionnaire designs, they are not suited for the collection of social interactions, including customer interactions. Behavioral research into social interactions can either be conducted in interactive or non-interactive settings~\cite{Arechar2018}. In non-interactive designs, participants are presented with a well-defined task description in which they react to a given stimulus in a specific social setup. Participants can complete the experiment individually and in parallel with others. Moreover, non-interactive studies can be split up into several rounds, such that the outputs of a prior phase can be given as input stimuli to participants in the next phase, thus imitating real-time interactions. In interactive studies, participants are matched with other individuals and engage in live repeated interactions.

Researchers in economics, social sciences or psychology frequently conduct economic games to model human behavior and decision-making in different types of social interactions~\cite{Thielmann2021}. As the goal is to test specific theories or detect correlations in participants' behavior, the communicative aspects of interactions are often not considered and potential variation in replies is removed by giving participants templates~\cite{Abeler2019,Agranov2024}. Other studies that focus more on the utterances within an interaction present participants with some context and a specific utterance, prompting them to respond using a predetermined strategy, such as reappraisal categories~\cite{Morris2015} or coping mechanisms~\cite{enri_emnlp2024}. Whereas Morris et al.~\cite{Morris2015} trained their participants to use a specific reappraisal or ‘therapeutic' technique, Troiano et al.~\cite{enri_emnlp2024} phrased their generation task as a role-playing game in which participants had to act out the description of a given character representing a specific coping strategy. Role-playing as a technique to prime human subjects for data generation is also used in live interactive studies on, e.g., cyberbullying~\cite{VanDenBroeck2014,Emmery2021} or applications of human-centered human--robot interaction~\cite{Dautenhahn2007}, including customer interactions~\cite{Song2022}. In some cases, the Wizard of Oz technique is considered a special case of role-playing in which a human wizard either partially or fully mimics the behavior of a computer system that appears to be autonomous to participants. 

\subsubsection{The Wizard of Oz Technique}
Even though similar experimenter-in-the-loop studies had already been conducted, the term \textit{Wizard of Oz} (WOZ) was conceptually introduced around 1980 by J.F. Kelley in the context of his PhD dissertation on natural-language user interfaces~\cite{Kelley1984,Green1985}. The term refers to a fictional character of Baum's \cite{Baum1900} children's novel. In that story, the Wizard is portrayed as the incredibly powerful ruler of the Land of Oz, but by the end of the story, it is revealed that the Wizard is in fact an ordinary conman who hides behind a screen and deceives everyone with magic tricks. As a research paradigm, the WOZ technique is used when a human experimenter or \textit{wizard} simulates a computer's behavior during interactions. This behind-the-scenes human manipulation can either be performed with the participant's prior knowledge or involve a low level of deception to encourage natural reactions. Moreover, the level of control an experimenter exercises varies between providing all functionalities to operating only a single component of an otherwise autonomous system. 


Over the years, the WOZ technique has been used across different research fields for the iterative development and evaluation of computer systems. One primary distinction in WOZ experiments lies in the communication modalities used. Fraser \& Gilbert \cite{Fraser1991} differentiate between WOZ simulations that do not involve natural language and those that do. WOZ experiments without natural language focus on alternative communication modalities such as gestures to execute actions~\cite{Connell2013}, body language to express affect~\cite{Andersson2002}, screen touches or mouse clicks to select items~\cite{Linnell2012}, or haptic feedback to warn users~\cite{Rector2018}.
Example applications of this type of WOZ simulations range from mobile game prototyping~\cite{Hoysniemi2004} to the development of autonomous driving technology~\cite{Wang2017}. 

Most WOZ experiments, however, involve natural language, either in the form of speech or typed text. These experiments can be further categorized into settings where only one of the participants uses natural language -- either the subject~\cite{Hauptmann1989} or wizard~\cite{Polavcek2012} -- and settings in which both the wizard and subject use it~\cite{Tanaka2012}. Initially, the technique was introduced to test the viability and usability of natural language applications, often involving speech~\cite{Gould1983}, and to inform the iterative prototyping of such systems~\cite{Kelley1984,Gould1985}. While the WOZ technique remains popular for these purposes, researchers have also used it to examine the behavior and language of human subjects interacting with applications. Dahlbäck et al.~\cite{Dahlback1993} argue that man-machine communication differs from human discourse, as humans often skip obvious reasoning steps and rely on background knowledge. Additionally, Leiser~\cite{Leiser1989} and Maulsby et al.~\cite{Maulsby1993} found that humans unconsciously mirror the speech patterns of the dialogue systems they interact with -- a phenomenon known as \textit{convergence}, where speakers adjust their speech patterns to match those of their conversation partners. By invoking this phenomenon during human-machine interaction, conversations can be kept within the range of the system's capabilities, ensuring a smooth flow.

\subsubsection{WOZ for Social Interaction and Emotion Datasets}

With technological advancements and the growing scope of its applications, the WOZ paradigm has also been embraced by researchers in social human-computer interaction. Its use spans various applications, from developing an attentive listening robot~\cite{inoue-etal-2020} to examining emotional versus rational persuasion strategies in computer-mediated dialogue~\cite{Adler2016}, as well as investigating the capabilities and limitations of LLMs as wizards~\cite{Fang2024}. In the realm of emotion analysis, the WOZ paradigm has contributed to the creation of several resources. One such resource is called deLearyous~\cite{Vaassen2012}, a small Dutch (Flemish) dataset that contains 11 written dialogues grounded in a single scenario. The corpus is annotated with the Interpersonal Circumplex~\cite{Leary1957}, which locates interpersonal behavior along two axes, dominance and affiliation. While this framework indirectly reflects affective states (e.g., dominant–hostile behavior may co-occur with anger) and is used by the authors as a proxy for emotion labels, this proxy is not ideal: the model targets behavior rather than affect, producing indirect, many-to-one, context-dependent mappings and increasing misclassification risk, especially when behavior is situationally driven rather than emotion-driven.

A second resource, the Distress Analysis Interview Corpus (DAIC)~\cite{gratch-etal-2014-distress}, is a multimodal dataset containing semi-structured clinical interviews designed to assess psychological distress conditions such as anxiety, depression, and post-traumatic stress disorder. It includes interactions with human interviewers, a human-operated virtual agent, and a fully autonomous system, providing a rich variety of conversational contexts and modalities to analyze distress signals. The dataset comprises audio and video recordings, extensive questionnaire responses, and annotations for verbal and non-verbal features. Participants include both distressed and non-distressed individuals, providing a broad perspective on responses across different mental health states. Although highly valuable, the DAIC corpus is specifically annotated for distress disorders rather than general emotional states.

Finally, EmoWOZ~\cite{feng-etal-2022-emowoz} is a large-scale corpus of task-oriented dialogues annotated for emotions. It includes user utterances from the MultiWOZ corpus~\cite{Budzianowski2018}, which features dialogues in various domains such as booking hotels and finding restaurants, along with an additional set of dialogues where humans interacted with an autonomous system. The emotion annotation scheme of EmoWOZ consists of six non-neutral categories and one neutral category. Labels are assigned to an utterance based on its valence, elicitors (user, operator, event), and user conduct (polite/impolite). The static categorization does not entirely account, however, for the nuances and context-dependent nature of human emotions, leading to potential mismatches. For instance, a user's polite expression of frustration might be labeled as \textit{apologetic}, even though the primary emotion being conveyed is frustration. Additionally, the inclusion of labels like \textit{apologetic} and \textit{abusive}, which describe behaviors rather than emotions, can cause confusion. Finally, the scheme's predefined categories may not cover the full spectrum of human emotions, potentially excluding or misrepresenting certain expressions. Our corpus, \corpus, seeks to overcome these limitations by implementing a more flexible annotation scheme and by actively steering the valence trajectory of conversations to elicit emotions in human--robot interaction---a targeted approach not adopted in EmoWOZ, where the authors relied on pre-existing data. Despite its name, \corpus~is a fully new dataset, independent of EmoWOZ and MultiWOZ in terms of data.



\section{The Corpus \corpus}\label{sec3:data_collect}
This section details the construction and contents of \corpus, focusing on experimental design, annotation protocol, and release specifics not covered in the introduction. We first describe the controlled Wizard of Oz data collection and operator strategy design (Section~\ref{sec3:method}). We then present the annotation pipeline for emotions (categorical labels and valence–arousal–dominance scores), operator response strategies, translation from Dutch (Flemish) to English, and anonymization procedures (Section~\ref{sec3:data_cleaning}). Finally, we summarize corpus statistics and availability. The dataset \corpus~is publicly released in both Dutch (Flemish) and English at \url{https://lt3.ugent.be/resources/emowoz-cs-en-nl/}.

\subsection{WOZ Experimental Design}\label{sec3:method} 
\begin{figure*}[h!]
\centering
\includegraphics[width=\textwidth]
{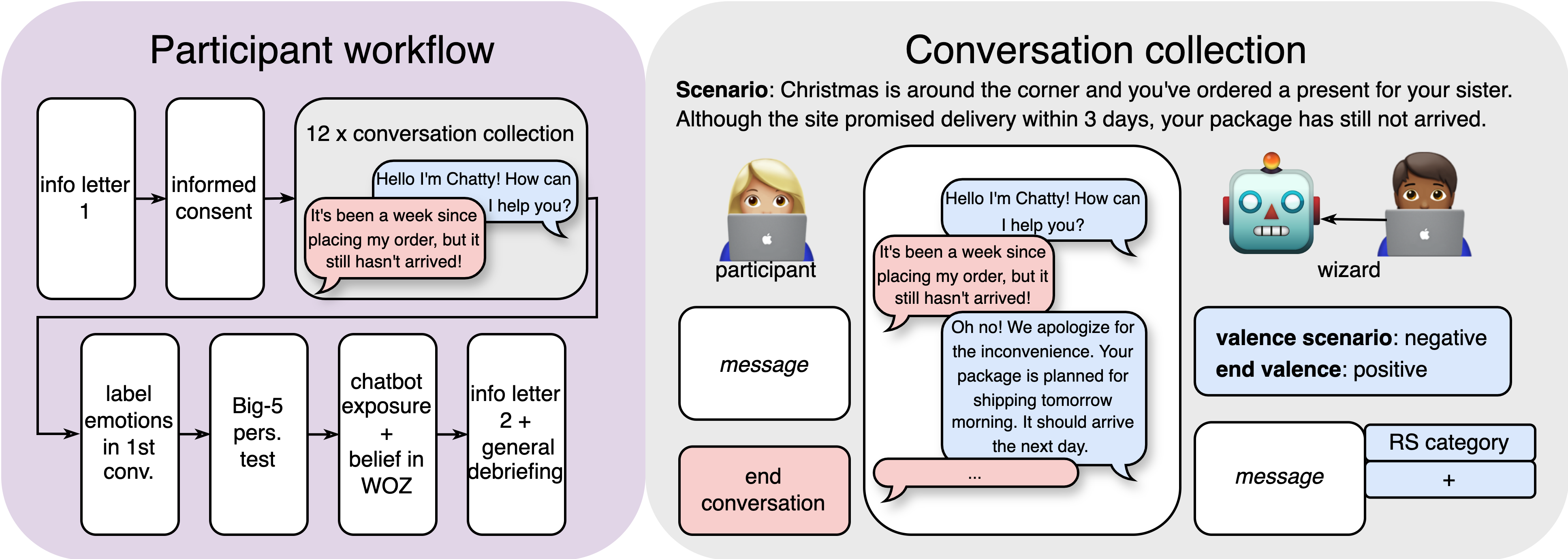}
\caption{WOZ experiment methodology. Purple box: full participant workflow, including the conversation collection, emotion labeling, and profiling questionnaires. Gray box: conversation collection with participants' unique functionalities in red and wizards' unique functionalities in blue. Both interlocutors read the scenario and can type messages. Wizards must label the response strategies (RS category) in their message before sending it, unlike participants. Participants decide about ending the interaction; wizards receive an emotional valence toward which to steer the conversation (end valence).}
\Description{The figure shows a two-part schematic of the Wizard of OZ (WOZ) experiment workflow. On the left, a purple boxed flow outlines the full participant journey: consenting and reading a scenario, engaging in a chat conversation, then completing emotion labeling and profiling questionnaires. On the right, a gray boxed diagram zooms into the chat collection phase, contrasting participant and wizard roles. Both sides display shared elements—each interlocutor reads the same scenario and can type messages—but participant-only actions are highlighted in red and wizard-only actions in blue. The participant has the ability to decide when to end the interaction. The wizard, who controls the agent side, receives an assigned target emotional valence to guide the conversation toward and must select a response strategy category for each message before sending.}
\label{fig:woz_method}
\end{figure*}

To collect our own corpus of customer interactions, we conducted a WOZ experiment in which human subjects engaged in written chat conversations with what they are told is an autonomous agent, while the agent is in reality fully controlled by a student worker. Our experimental setup was informed by a pilot study~\cite{Labat2022a}. The final setup of our current experiment is illustrated in Figure~\ref{fig:woz_method} and described in what follows. The experiment was submitted to and approved by the Ethics Committee of the Faculty of Arts and Philosophy at Ghent University.

\subsubsection{Recruitment of Wizards and Subjects} \label{sec3:recruit_wiz}
We recruited eight student workers from the Faculty of Arts and Philosophy at Ghent University as wizards. Each wizard assumed the role of a conversational agent, functioning as a customer service agent. Student workers were remunerated per hour (in accordance with the university's salary scales) and were expected to find participants in their social network. Before beginning their roles, wizards underwent a one-hour training session. During this session, they were thoroughly briefed on their tasks and how the in-house online application used in the experiment functioned.

In total, 179 persons were recruited by wizards and took part in our study. Potential candidates had to be at least 18 years old, speak Dutch as their mother tongue, and have a computer with stable internet connection available. After completing the experiment, participants received a movie ticket as compensation for their time and effort.

\subsubsection{Information Letters and Informed Consent}
Before the start of the experiment, we presented participants with an initial information letter. After reading the letter, participants had to first give their informed consent before they could participate in the experiment. To ensure natural and spontaneous reactions, the letter explained that participants would be chatting with various autonomous conversational agents, 
while in reality, these agents were fully controlled by a human operator. For the same reason, we did not disclose that we would study the emotions expressed in their replies. Both pieces of information were revealed at the end of the experiment through a second information letter. With this new information, participants were explicitly informed that they could still withdraw their consent and remove their data from the study up to seven days after their participation, after which their data would be anonymized. 

\subsubsection{Conversation Collection}\label{sec3:experiment_details}
Each participant engaged in 12 different conversations with a wizard. Participants took on the role of customers, while wizards acted as customer service operators. Both interlocutors read the same scenario to use as the basis for their conversation. In our graphical user interface, only participants had the functionality to end conversations. In contrast to participants, wizards received information on the valence trajectory and were required to label their messages with response strategies before sending them. 
The full conversation collection process lasted roughly one and a half hours per participant. While Figure~\ref{fig:woz_method} contains a schematic overview of the conversation collection methodology, an actual conversation from \corpus~is portrayed in Figure~\ref{fig:ex_conv} of Appendix~\ref{app:examples}. The remainder of this section zooms in on the scenarios, valence trajectories, and operator response strategies.

\paragraph{Scenarios to Ground Interactions.} Each conversation was grounded in a description of an event (henceforth \textit{scenario}) linked to either a negative or neutral valence. We did not include positive valence, as subjects of our pilot study had indicated that it felt unnatural to contact a conversational agent to discuss positive events in the realm of customer service. The proposed scenarios concern four economic industries with customer-facing operations: commercial aviation, e-commerce, online travel agencies, and telecommunication. 

A total of 208 distinct scenarios were manually crafted, with 52 scenarios per economic sector. Each set of 52 scenarios was further split, featuring an even distribution of negative and neutral valence per economic sector. The resulting scenarios were randomly distributed across participants, ensuring that each participant received 12 unique scenarios.\footnote{Three participants inadvertently received the same scenario twice. Their IDs, as recorded in the dataset, are 10, 14, and 15. The occurrence does not affect the analysis, experiments, or data quality; it is noted here for transparency.} For illustration, Table~\ref{tab:scenarios} in Appendix~\ref{app:examples} contains examples of scenarios used in the experiment.

\paragraph{Valence Trajectories.}
Valence trajectories provide wizards with a guiding framework for steering the conversation. Each valence trajectory is a pair of valence values: the first represents the scenario's initial valence (valence scenario), while the second indicates the target valence in which the wizard aims to conclude the conversation (end valence). These trajectories are visible to wizards in our graphical user interface but remain hidden from participants. Conversations can end with one of three valence types: negative, neutral, or positive. We aimed to distribute these three types equally across scenarios. However, minor discrepancies remain because the number of scenario occurrences (either 10 or 11 times) is not evenly divisible by the three valence types. Additionally, we aimed to ensure a uniform distribution of end valence across participants.\footnote{Due to an oversight, six participants received an uneven distribution of end valence types. Their IDs are 2, 3, 14, 50, 108, and 120. It does not impact the analysis, experiments, or data quality; it is mentioned for transparency.}

Even though the goal of valence trajectories is to induce emotion shifts during the interaction, in some cases the end valence will be identical to the valence of the scenario in which the conversation starts. We argue that such instances are necessary, as they also occur in real-life, non-simulated customer interactions. Furthermore, whereas real-life human operators will almost always attempt to increase the valence of customers during complaint handling, our dataset also contains instances in which wizards have to navigate interactions toward a negative valence. The addition of these somewhat atypical trajectories is motivated by the fact that autonomous conversational agents with non-deterministic response generation may occasionally make undesirable mistakes. A dataset of such undesirable replies will improve a system's capabilities in detecting them and subsequently preventing the generation of similar messages.

\paragraph{Operator Response Strategies.}\label{sec3:annot_rs}

Rather than using fixed templates, we chose to give wizards the flexibility to create their own responses to steer the conversation along a valence trajectory. This approach enhanced text diversity and provided the necessary freedom for tailoring replies. To offer some low-level guidance and insight into semantically similar responses, we introduced a set of operator response strategies. Before sending a message, wizards were required to label their responses with the relevant strategies in a multilabel setup. Response strategies are organized into four groups of techniques, as shown in Table~\ref{tab:rs_strat_list}. Besides helping wizards, operator response strategies are also used to improve control over emotion elicitation, to identify effective and ineffective strategies, and to provide interpretable metadata on our corpus. Furthermore, the labels and texts can be leveraged to train AI systems in task-oriented dialogues and develop emotion-aware dialogue systems.

\begin{table}[h!]
\caption{Summary of response strategies used in the experiment.} 
\label{tab:rs_strat_list} 
\centering 
\small
\begin{tabular}{ll} 
\toprule 
Technique & Response strategies\\ 
\midrule 
Affective & Apology, cheerfulness, empathy, gratitude\\
Objective & Explanation, help offline, help online, request action, request information\\ 
Suboptimal & Inappropriate response, irony, miscomprehension, non-collaborative response\\
Alternative & Other\\
\bottomrule 
\end{tabular}
\end{table}

The taxonomy in Table~\ref{tab:rs_strat_list} is a revised version of the original annotation scheme~\cite{Labat2020} used in our previous studies~\cite{Labat2022a,Labat2024}. Updates include the introduction of the suboptimal label group, which was previously categorized under \textit{other}, as well as the addition of the objective techniques \textit{help online} and \textit{request action}. Further details on the meaning of the different response strategies, along with examples, are provided in Table~\ref{tab:rs_annot} in Appendix~\ref{app:annotation}. 

\subsubsection{Emotion Labeling by Participants} \label{sec3:part_emo_annot}
After collecting the conversations, participants were asked to annotate their own messages in the first conversation with emotion labels. We adopted a taxonomy of 11 emotions 
(\textit{admiration}, \textit{anger}, \textit{annoyance}, \textit{confusion}, \textit{desire}, \textit{disappointment}, \textit{fear}, \textit{gratitude}, \textit{joy}, \textit{relief}, and a fallback category \textit{emotion label not found}), together with a \textit{neutral} category (see Table~\ref{tab:emo_annot} in Appendix~\ref{app:annotation} for details). The rationale and refinement of this taxonomy are described in Section~\ref{sec3:annot_emo}. Unlike the external annotators (see Section~\ref{sec3:annot_emo}), participants did not receive specific training for emotion labeling. Nevertheless, these self-reports are valuable as they capture participants' subjective emotional experiences, providing a basis for later comparison with externally assigned annotations.

\subsubsection{Participant Questionnaires} \label{sec3:part_question}
One major benefit of studies involving human subjects is the opportunity to gather supplementary profiling data on participants. This information helps analyze distribution variations among individuals and can potentially enhance the performance of predictive algorithms. In this study, we collected demographic variables (age, gender), personality traits, participants' familiarity with the WOZ setup, and their prior experience with chatbots, as summarized in Table~\ref{tab:participant_info}.

\begin{table}[h!]
\caption{Summary of participant information in the \corpus~study.}
\label{tab:participant_info}
\centering
\small
\begin{tabular}{ll}
\toprule
Participant Information     & Variables\\ 
\midrule
Demographics                & Age, Gender\\
Personality Traits          & Agreeableness, Conscientiousness, Extraversion, Neuroticism, Openness\\
Familiarity with WOZ        & Awareness rated on a 4-point Likert scale\\
Prior Chatbot Experience    & Participants asked about their prior chatbot history\\
\bottomrule
\end{tabular}
\end{table}

\paragraph{Big-Five Personality Test.}
We adopted a well-established five-factor model to measure personality along five traits: agreeableness, conscientiousness, extraversion, neuroticism, and openness~\cite{Goldberg1992}. The instrument we used is the IPIP-NEO-120 personality test~\cite{Johnson2014}, a 120-item questionnaire designed to assess these five traits using a comprehensive set of self-report items. Each item of the test consists of a statement related to a specific trait, and participants are asked to rate how accurately the statement describes themselves on a 5-point Likert scale. To safeguard participant privacy, we integrated a Dutch version of the test into our in-house online application. This approach ensured that no third-party providers were involved. In the \corpus~corpus, we report an aggregated score for each personality trait. These scores, ranging from 24 to 120, indicate the strength of the trait, with higher scores reflecting a greater presence of that particular trait.

\paragraph{Belief in WOZ and Prior Chatbot Experience.} We asked participants whether they had realized they were interacting with a human wizard rather than an autonomous chatbot. They rated their awareness on a 4-point Likert scale, with 1 indicating complete unawareness and 4 indicating high confidence in recognizing the WOZ setup. Participants selecting 3 or 4 were further prompted to describe the cues that informed their judgment. In a follow-up yes/no question, we asked participants about their prior experience with conversational agents.

\subsection{Data Preparation and Preprocessing} 
\label{sec3:data_cleaning}
After the participatory phase, we cleaned the raw data. This data preparation phase involved external annotators tagging participant messages for emotions (Section~\ref{sec3:annot_emo}), translators converting the Dutch conversations to English (Section~\ref{sec3:translate}), and both automatic and manual steps to remove personal information on participants from texts (Section~\ref{sec3:privacy}).

\subsubsection{Emotion Annotation}\label{sec3:annot_emo}

Whereas the participant labels reflect first-person perspectives, external annotators were tasked with labeling all conversations using the same set of 11 emotion categories plus \textit{neutral} (see Section~\ref{sec3:part_emo_annot}). In addition, they evaluated each message along three continuous dimensions (valence, arousal, and dominance) on a 5-point scale. For this process, we recruited two trained student workers from the Faculty of Arts and Philosophy at Ghent University, both of whom had previously served as wizards and were compensated hourly in line with the university's salary scales.

The taxonomy of emotion labels was informed by insights from our pilot study, which initially included a broader set of 16 labels~\cite{Labat2022a}. To improve reliability and interpretability, we refined this set by removing labels that were infrequent, excluding labels with low inter-annotator agreement, and collapsing semantically overlapping categories (e.g., merging \textit{sadness} into \textit{disappointment}). Furthermore, we introduced the new label \textit{relief} after our pilot study revealed its absence and we noticed that no closely related alternative existed within the broader set of 16 labels. To ensure comprehensive coverage of all possible emotional expressions, we also added the residual category \textit{emotion label not found}, which functions as a fallback for annotators when a message conveys an emotion that does not fit any of the predefined categories. Its inclusion allows for flexibility, ensuring unique or rare emotions are accurately represented without compromising the refined taxonomy. Table~\ref{tab:emo_annot} of Appendix~\ref{app:annotation} provides an overview of our taxonomy and label definitions as presented to annotators.

\subsubsection{Creation of Parallel Corpus in English}\label{sec3:translate}
Since the WOZ experiment was conducted in Dutch, we translated the collected text messages from Dutch to English. The resulting parallel corpus enables broader accessibility and supports research in machine translation and multilingual NLP applications, particularly those focused on textual interactions. To generate a parallel corpus, we first used DeepL\footnote{\url{https://www.deepl.com/en/translator/}}  to automatically translate the Dutch text messages into American English. The original and translated messages were then aligned in a TMX file using YouAlign\footnote{\url{https://youalign.com/}}. This file was subsequently imported as a translation memory into MateCat\footnote{\url{https://www.matecat.com/}}, an open-source computer-assisted translation (CAT) tool. 

To ensure data quality, three student workers from Ghent University post-edited the automatically translated texts in MateCat. Two of those students were pursuing a Master's in Translation, while one was enrolled in a Master's in Multilingual Communication. They were selected based on their translation speed and accuracy in a test task, and were paid hourly according to the university's salary scales. Before starting, the post-editors attended a training session on the corpus and experimental setup. They were instructed to use light post-editing, focusing on speed and correcting only critical issues such as typos, grammatical errors, mistranslations, and semantic mistakes. A key priority was ensuring that the underlying emotional meaning of the source messages was accurately conveyed in the target language.

\subsubsection{Data Anonymization}\label{sec3:privacy}
Participants were asked to interact with the automatic chatbot as they normally would during a customer service interaction. To protect their privacy, we repeatedly instructed them not to share any personal information during the experiment. Additionally, we gave them a fictional client profile with personal data to use if prompted for this information by the operator. Despite these precautions, some participants inadvertently shared data not linked to the fictional profile, and it was unclear whether this data was their own or fabricated. To prevent any accidental sharing of personal information, we automatically replaced strings that could potentially be used to identify persons with general placeholders and manually reviewed and corrected the dataset for any missed instances. We also replaced the names of existing companies that occurred in scenario descriptions and messages during the experiment. For the scenarios, we chose companies active in Flanders to help participants relate to the fictive scenarios. In the finalized corpus, these specific names were replaced with general placeholders. Details on the placeholders and their corresponding items are in Appendix~\ref{app:anonym}.

\section{Analysis of \corpus}\label{sec4:analysis}
Following the creation of \corpus, we aim to gain a deeper understanding of the dataset's composition and interactional dynamics. This section is structured into four parts. First, we provide descriptive statistics on the collected conversations in Section~\ref{sec4:corpus}. Next, we examine our participants' profiles in Section~\ref{sec4:profiles}, including demographics, personality traits, their knowledge about the WOZ setup, and prior exposure to chatbot technologies. In Section~\ref{sec4:emotion}, we explore the distribution of emotions and assess the reliability of annotations through inter-annotator agreement, including comparisons between self-reported and third-party labels. Finally, we analyze the conversational dynamics in Section~\ref{sec4:conv_dynamics}, focusing on the strategies applied by operators and how these relate to the emotional trajectories of the interactions. 

\subsection{Descriptive Statistics on Conversations} \label{sec4:corpus} 
The resulting \corpus~corpus consists of a total of 2,148 conversations, which were collected from 179 participants who each engaged in 12 different dialogues with a wizard. These conversations collectively contain 23,706 individual messages. On average, a conversation contains 11.04 messages, with a standard deviation of 5.71, indicating some variation in conversation length. The shortest conversation in the dataset includes just 3 messages, while the longest has 47 messages.

\begin{figure*}[ht!]
\begin{center}
\includegraphics[width=\textwidth]
{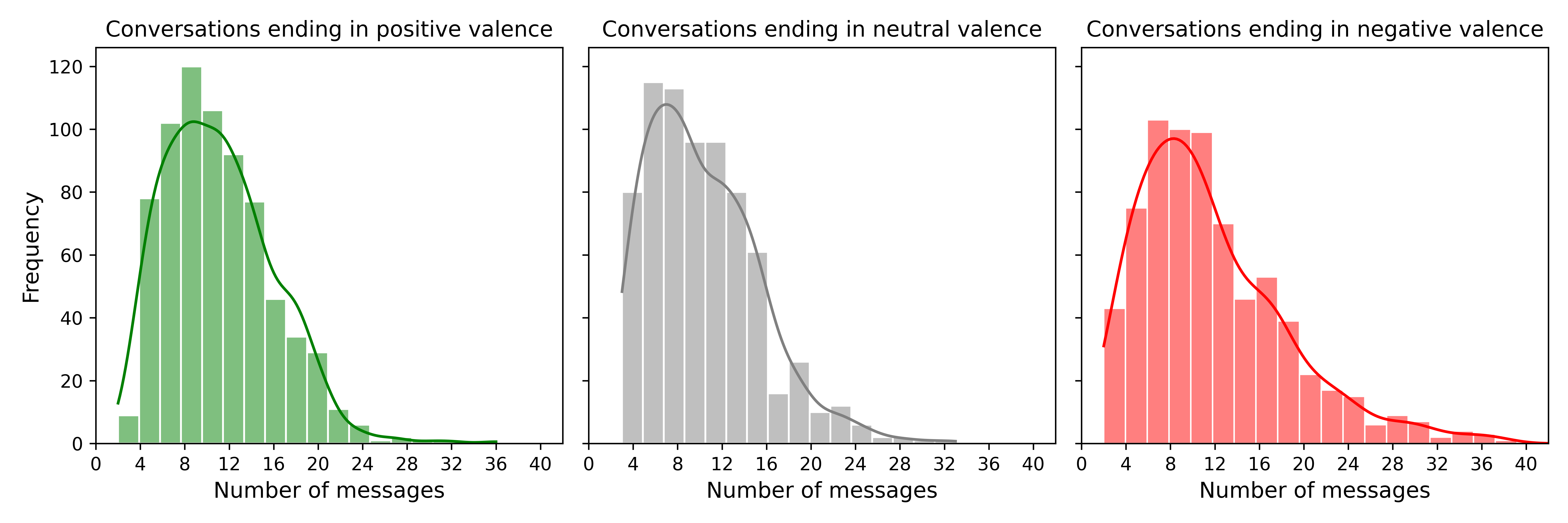}
\end{center}
\caption{Histograms of message counts by prescribed final valence categories.}
\label{fig:hist_n_messages_per_val}
\Description{Three side-by-side histograms show distributions of conversation lengths (number of messages) split by prescribed end valence: positive, neutral, and negative. Each panel has the same horizontal axis for message count and a vertical axis for absolute number of conversations, allowing direct comparison across categories with similar sample sizes. The positive panel has a peak around 8–10 messages with a moderate right tail. The neutral panel peaks slightly lower, around 7–9 messages, and closely resembles the positive distribution. The negative panel is more dispersed, with a broader spread and a pronounced long tail that includes bins beyond 40 messages, indicating occasional very long conversations. All three distributions are right-skewed, with the negative category exhibiting the greatest variability. Bin bars are plotted without normalization, and axes, legends, and titles indicate counts, conversation length, and valence category.}
\end{figure*}

To investigate how the intended emotional outcome influences conversation length, we analyzed the number of messages per conversation based on the \textit{prescribed end valence} -- i.e., the emotional tone wizards were instructed to guide the conversation toward. It is important to note that this target valence does not always align with the participant's actual emotional state at the end of the interaction. Figure~\ref{fig:hist_n_messages_per_val} presents three histograms comparing conversation lengths across positive, neutral, and negative target valence categories. Since the number of conversations is nearly identical across these categories (716 positively ending, 717 neutrally ending, 715 negatively ending), we report absolute message counts rather than normalized frequencies. The histograms show right-skewed lengths across all valence categories, with the clearest separation in the long tail. Conversations with positive target valence cluster around 8--10 messages, neutral ones peak slightly lower around 7--9 messages, and overall these two distributions look quite similar. In contrast, conversations with negative target valence have the greatest variability in length, with a heavy tail extending beyond 40 messages. This pattern suggests that negative interactions often persist, likely due to conflict or difficulty reaching resolution, although some negative exchanges are also cut short. Overall conversation length seems to function as a proxy for unresolved complexity: brief exchanges tend to signal resolution, whereas prolonged back-and-forth increasingly correlates with negative outcomes.

To better understand the linguistic composition of our dataset, we analyzed the token distributions across Dutch and English using the spaCy library\footnote{\url{https://spacy.io/}}. Emojis\footnote{\url{https://pypi.org/project/emoji/}} and anonymized placeholders (see Appendix~\ref{app:anonym}) were treated as single tokens. The Dutch corpus contains 284,282 tokens (247,833 excluding punctuation), while its parallel English version includes 294,418 tokens (255,633). Of the total messages, 10,858 were written by participants and 12,848 by wizards. Message lengths are comparable across roles: participant messages average 12.26 tokens in English (10.79 excluding punctuation) and 11.88 in Dutch (10.52), while wizard messages are slightly longer, averaging 12.56 tokens in English (10.78 excl. punctuation) and 12.09 in Dutch (10.40). This suggests that while wizards contribute slightly more messages, both roles seem to be similarly verbose in their individual turns.


\subsection{Participant Profiles} \label{sec4:profiles}
This section outlines the characteristics of the participant sample, including demographics and personality traits. It also explores participants' knowledge about the WOZ setup and their prior experience with chatbots. 

\subsubsection{Demographics and Personality Traits}
Of the 179 participants, 74.9\% (n=134) identified as women and 25.1\% (n=45) as men. The age distribution of participants is concentrated in the early twenties, with individuals aged 18 to 26 accounting for over 83\% of the total sample. The most frequently reported ages are 21 (n=43, 24.0\%), 22 (n=42, 23.5\%), and 20 (n=18, 10.1\%). Representation declines sharply beyond the age of 26, with only isolated cases at older ages, including a small number of participants aged 51 (n=4), 47 (n=3), and 55 (n=2). While the youngest participant was 18 years old, the oldest was 83. The skew in age distribution reflects our recruitment strategy (see Section~\ref{sec3:recruit_wiz}): eight student workers served as wizards and recruited participants from their social networks, which are predominantly student-oriented, leading to an overrepresentation of ages 20–22.

\begin{figure*}[t!]
\begin{center}
\includegraphics[width=.92\textwidth]
{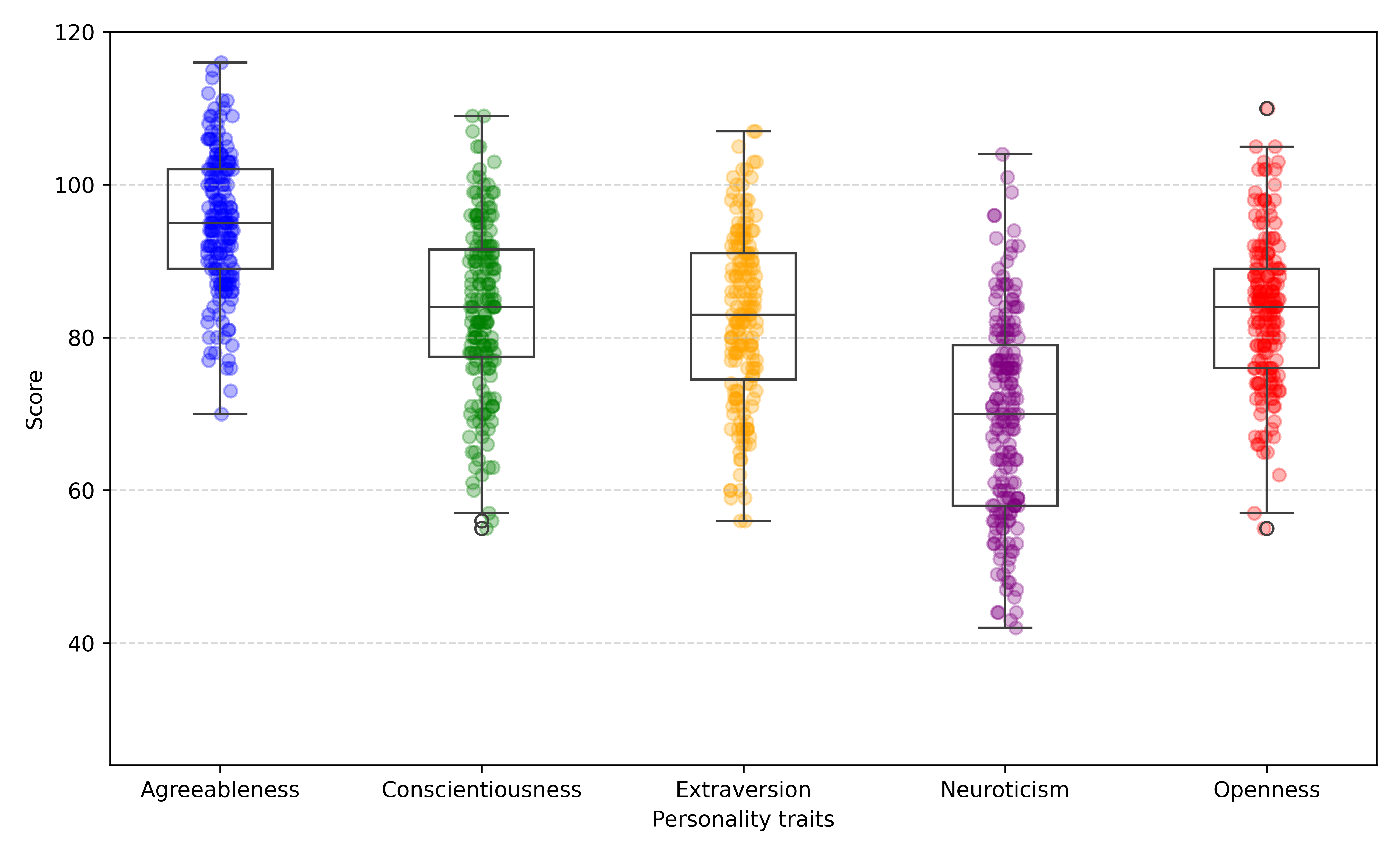}
\end{center}
\caption{Big Five personality traits distribution.}
\label{fig:boxplot_personality}
\Description{The figure shows a set of five side-by-side boxplots (overlaid with scatter) summarizing participants' scores (range 24–120) for agreeableness, conscientiousness, extraversion, neuroticism, and openness. Each boxplot displays the median line, interquartile range, whiskers, and individual outliers. Agreeableness has the highest median with few outliers, indicating clustered, higher scores. Conscientiousness, extraversion, and openness have medians around the low-to-mid 80s with moderate spread and few outliers. Neuroticism exhibits the widest spread with a lower median than most other traits and multiple low-end values, including points below 60, highlighting substantial variability. Axes include a horizontal score axis labeled from 24 to 120 and categorical labels for the five traits.}
\end{figure*}

In addition to providing age and gender information, participants completed a personality test (see Section~\ref{sec3:part_question} for details) with possible scores ranging from 24 to 120 for each trait. Figure~\ref{fig:boxplot_personality} presents the distribution of scores across the five personality traits: agreeableness, conscientiousness, extraversion, neuroticism, and openness. Median scores for conscientiousness, extraversion, and openness fall between 80 and 90. Neuroticism displays the greatest range in scores, with several participants scoring well below 60, indicating substantial variation for this trait within the sample. Conversely, agreeableness has the highest median score and the narrowest distribution, suggesting more consistency among participants for this trait.

\subsubsection{Assessing Familiarity with WOZ and Prior Chatbot Experience}
Most participants (viz. 44.1\%, n=79) reported that they did not know they had been interacting with a human. Another 29.6\% (n=53) had a slight suspicion but still leaned toward believing they were interacting with a chatbot. Meanwhile, 21.2\% (n=38) had a strong suspicion they were chatting with a human, and only 5.0\% (n=9) indicated they were certain of this. In a follow-up question to the 47 participants who either strongly suspected or knew they were interacting with a human, several indicators were identified that revealed the true nature of the setup. Response timing was frequently mentioned, with delays in replies and inconsistent response speeds raising doubts about automation. Linguistic cues such as emotional tone, sarcasm, humor, and even occasional spelling errors contributed to a more human-like impression. Participants also pointed to unexpected or inappropriate responses, empathetic or personal replies, and use of emojis, which they believed deviated from typical chatbot behavior. Prior experience with chatbots influenced perceptions as well, with some participants noting that the responses were less formal, more impulsive, or more contextually nuanced than they would expect from a chatbot. A few participants mentioned being accidentally informed by the human wizard or suspecting the setup due to the nature of the experiment.

Regarding prior experience with conversational agents, 77.7\% of participants (n=139) had interacted with one before, while 22.4\% (n=40) had not. This background likely influenced their ability to detect human-like behavior during the interaction. Moreover, the dataset was collected before the public release of ChatGPT, a development that has since shaped public expectations and familiarity with conversational AI.

\subsection{Emotion Annotation Analysis} \label{sec4:emotion}
We now present a detailed analysis of the emotion annotations collected in our study. We begin by examining the overall distribution of emotion labels across participant messages. Next, we assess the reliability of these annotations through inter-annotator agreement metrics, focusing on both categorical and dimensional emotion representations. Finally, we compare third-party annotations with participants' self-reported emotions to explore the alignment between perceived and experienced affect.

\subsubsection{Emotion Distribution}
Participant messages were annotated by one of two human annotators using both discrete emotion labels (in a multilabel setup) and 5-point ordinal ratings for valence, arousal, and dominance. The distribution of emotion labels is summarized in Table~\ref{tab:emotion-distribution}. Across the full dataset of 10,858 participant messages, the most frequent label was \textit{neutral}, accounting for nearly half of all messages (49.4\%, n=5,365). The high proportion of neutral messages reflects the task-oriented nature of customer service, where interactions typically prioritize efficiency and task completion over emotional expression. Even in our WOZ setup -- where participants role-played as customers -- the dialogue mirrored real-world service conventions. This corresponds to findings from the EmoTwiCS corpus~\cite{Labat2024}, which also reported a high frequency of objective messages in customer support on public social media channels.

\begin{table}[ht]
\caption{Distribution of emotion labels across 10,858 participant messages.}
\label{tab:emotion-distribution}
\centering
\small
\begin{tabular}{lrr}
\toprule
Emotion Label & Support (n) & Percentage (\%) \\
\midrule
Neutral         & 5,365 & 49.4 \\
Desire          & 2,008 & 18.5 \\
Gratitude       & 1,575 & 14.5 \\
Annoyance       & 820   & 7.6  \\
Anger           & 782   & 7.2  \\
Disappointment  & 416   & 3.8  \\
Joy             & 386   & 3.6  \\
Confusion       & 221   & 2.0  \\
Fear            & 128   & 1.2  \\
Admiration      & 33    & 0.3  \\
Relief          & 32    & 0.3  \\
\bottomrule
\end{tabular}
\end{table}

Among non-neutral messages, \textit{desire} (18.5\%, n=2,008) and \textit{gratitude} (14.5\%, n=1,575) were most prevalent, capturing customer intent and satisfaction, respectively. In contrast, \textit{annoyance} (7.6\%, n=820), \textit{anger} (7.2\%, n=782), and \textit{disappointment} (3.8\%, n=416) were also frequent, but signal friction or unmet expectations, thus pointing to an underlying dissatisfaction. \textit{Joy} (3.6\%, n=386) appeared with moderate frequency, reflecting moments of positive engagement other than expressions of \textit{gratitude}. Less frequent emotions were \textit{confusion} (2.0\%, n=221), \textit{fear} (1.2\%, n=128), while \textit{admiration} (0.3\%, n=33) and \textit{relief} (0.3\%, n=32) were rare. \textit{Admiration} was retained because it showed higher inter-annotator agreement than many other emotion categories in both EmoTwiCS~\cite{Labat2024} and our pilot study~\cite{Labat2022a}, despite its low frequency in the current dataset. This low occurrence likely reflects a design change, as explicitly positive scenarios were excluded after pilot study participants found such conversation starters unnatural. \textit{Relief} was added based on participant feedback during the pilot, where several users reported missing this label in the emotion taxonomy.

\begin{figure*}[ht!]
\begin{center}
\includegraphics[width=\textwidth]
{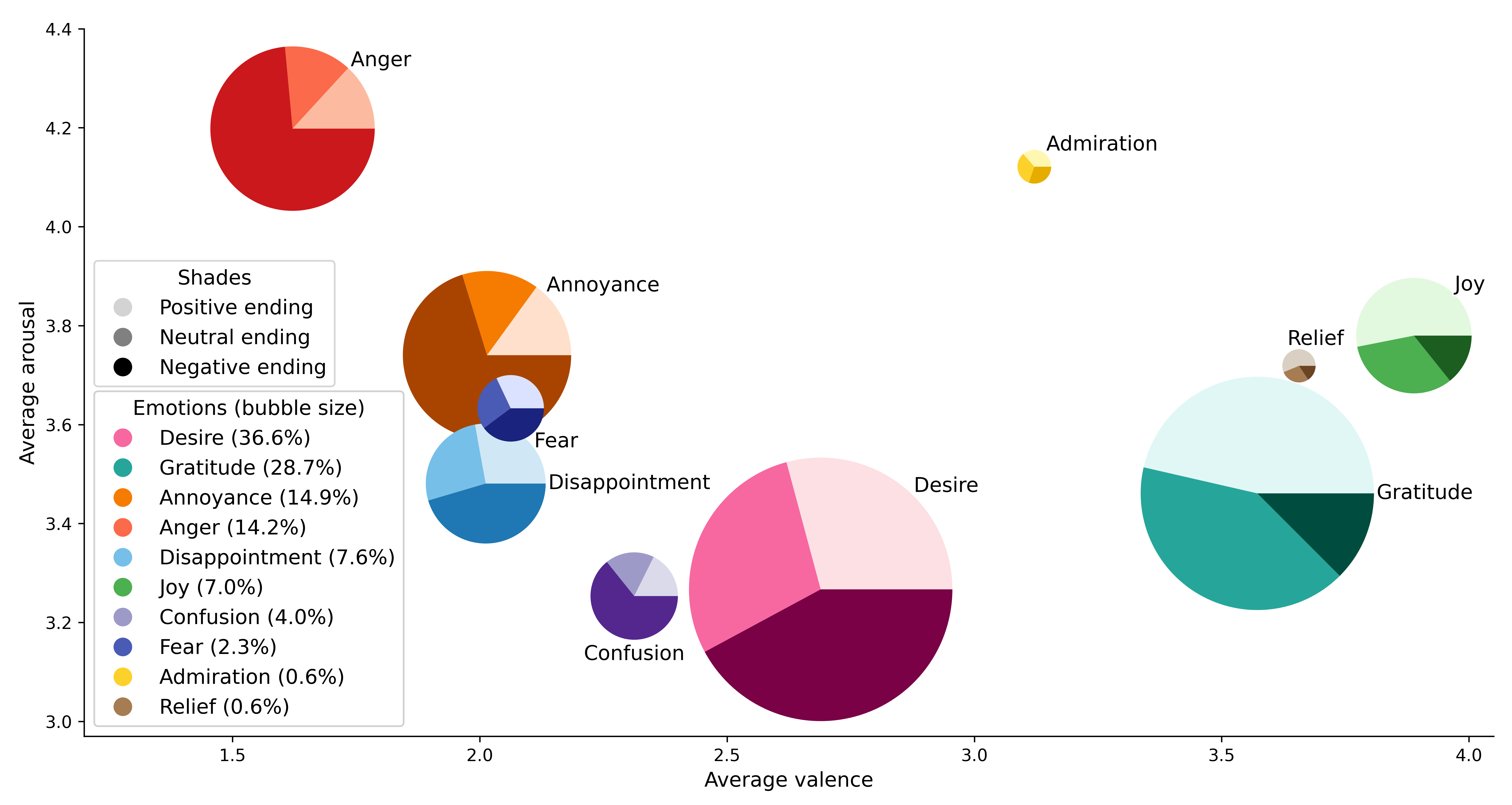}
\end{center}
\caption{Bubble pie chart showing the distribution of emotions in a two-dimensional affective space (valence × arousal). Each bubble represents an emotion, positioned by its average valence (horizontal axis) and average arousal (vertical axis). The size of each bubble reflects the relative frequency of that emotion among all messages labeled with an emotion (excluding neutral messages). Each bubble is divided into colored shades indicating the proportion of positive, neutral, and negative target valence outcomes for that emotion.}
\label{fig:emo_distrib}
\Description{The figure depicts a two-dimensional scatter of labeled bubbles placed in valence–arousal space, with valence increasing left to right and arousal increasing bottom to top. Each bubble corresponds to a specific emotion and varies in size by how often that emotion appears among non-neutral messages in the corpus. Gratitude, joy, and relief cluster on the right with moderate–high arousal; gratitude and joy are larger. Anger and annoyance are positioned further left (more negative valence) and at higher arousal than fear and disappointment. Confusion and desire lie slightly left of neutral with low arousal. Admiration appears upper-right with high arousal but only mildly positive valence and a small bubble. Each bubble is split into segments showing proportions of positive, neutral, and negative target endings (e.g., gratitude/joy/relief skew positive/neutral; anger/annoyance skew negative). Axes are labeled; emotion names appear near or inside bubbles.}
\end{figure*}

The affective distribution of emotions within a two-dimensional valence--arousal space is visualized in Figure~\ref{fig:emo_distrib}, where each emotion is represented as a bubble whose size reflects its frequency and whose color shades correspond to the target valence categories. Emotions such as \textit{gratitude}, \textit{joy}, and \textit{relief} reside in the high-valence region, while \textit{anger}, \textit{annoyance}, \textit{disappointment}, and \textit{fear} are clustered in the low-valence space. Positioned in the lower-valence region but less negative than the previously mentioned cluster, \textit{confusion} and \textit{desire} also stand out for their relatively low arousal levels. While commonly treated as emotions in NLP due to their lexical salience in text~\cite{Demszky2020}, psychological theory typically classifies \textit{confusion} as an affective-cognitive state and \textit{desire} as an affective-motivational one~\cite{Ellsworth2003,Berridge2018,Janssens2025}. Both share features with basic emotions, such as subjective experience and behavioral relevance, but they do not fully align with standard definitions of basic emotions. Their position in the valence--arousal space reflects this ambiguity, occupying a middle ground. Finally, \textit{admiration} is the second most arousing emotion, yet its average valence is only mildly positive -- possibly due to its relatively low frequency in the dataset, which may amplify the influence of atypical or ironic contexts.

Another pattern of interest concerns the colored shades within each bubble in Figure~\ref{fig:emo_distrib}, which represent the distribution of positive (light shade), neutral (medium shade), and negative (dark shade) target valence outcomes for each emotion. The proportions indicate how often a given emotion occurred in a conversational context where the wizard was instructed to steer toward a particular target valence. Emotions such as \textit{gratitude} (46.4\% positive, 41.1\% neutral), \textit{joy} (53.1\% positive, 32.6\% neutral), and \textit{relief} (56.3\% positive, 28.1\% neutral) are predominantly linked to conversations with a positive target valence. In contrast, negative emotions such as \textit{anger} (73.5\% negative), \textit{annoyance} (70.2\% negative), and \textit{disappointment} (45.4\% negative) are primarily associated with conversations with negative valence targets. \textit{Fear} and \textit{desire} show more mixed distributions, with considerable proportions of both negative (39.8\% and 42.1\%, resp.) and non-negative (60.2\% and 57.9\%, resp.) target outcomes. Interestingly, although \textit{confusion} carries only a mildly negative average valence, it appears most frequently in conversations with a negative target valence (64.3\% negative). This finding may indicate that \textit{confusion} arises as a byproduct of the suboptimal wizard response strategies (e.g., \textit{miscomprehension}, \textit{non-collaborative responses}) used to steer the emotional trajectory of the conversation. Finally, \textit{admiration}, despite its mildly positive valence, is evenly distributed across positive (36.4\%), neutral (33.3\%), and negative (30.3\%) target outcomes, possibly due to variability in its contextual use. Together, these distributions suggest that emotions serve distinct functional roles within emotionally guided interactions.

Beyond the placement and target valence distribution of individual emotions, broader structural patterns also emerge in the valence--arousal space. Figure~\ref{fig:emo_distrib} reveals a V-shaped curve across the valence spectrum, with arousal increasing at both ends and the lowest arousal occurring near neutral valence. This pattern aligns with prior findings at the nomothetic level: on average, people experience greater arousal when they feel strongly negative or positive emotions compared to neutral ones~\cite{Warriner2013,Labat2024}. However, Kuppens et al.~\cite{Kuppens2013} emphasize that the valence--arousal relationship varies substantially across individuals and contexts, implying that such trends may not be universal. In our dataset, we also observe that average arousal ratings never fall below the midpoint of the 5-point scale. This could reflect the purposeful nature of the interactions: as participants were actively engaged in addressing issues or goals, their affective states may have been characterized by a sustained baseline of emotional activation.

\subsubsection{Inter-Annotator Agreement}
Having explored broader affective patterns in the previous section, this section focuses on the reliability of the underlying emotion annotations that support these findings. To evaluate the consistency of emotion annotations, we computed inter-annotator agreement on a subset of the dataset consisting of 179 conversations (one conversation per participant), encompassing a total of 838 messages. The annotations analyzed in this section were provided exclusively by three external third-party annotators. While the full dataset is annotated by two of these three annotators (see Section~\ref{sec3:annot_emo}), a third annotator -- namely, the main author of this paper -- was added for this subset to enable a more robust assessment of inter-rater agreement. To assess the reliability of our emotion annotations, we employ Krippendorff's alpha~\cite{Krippendorff2004} and Fleiss' kappa~\cite{LandisKoch1977}, two complementary metrics for inter-rater agreement. Fleiss' $\kappa$ is used in Table~\ref{tab:emotion_agreement_sorted} to measure agreement on mutually exclusive emotion categories, where each message is assigned a single label (i.e., the emotion is either present or absent in a message). Krippendorff's $\alpha$, by contrast, is applied more broadly: it is used in this context as well, but also to assess agreement in multilabel annotation setups and for dimensional ratings.

\begin{table}[h]
\caption{Inter-annotator agreement on emotion annotation.}
\label{tab:iaa_general}
\centering
\small
\begin{tabular}{llc}
\toprule
Annotation task & Metric & Score \\
\midrule
Emotion labels & Krippendorff's $\alpha$ (Jaccard distance) & 0.6090 \\
Valence &  Krippendorff's $\alpha$ (absolute distance) & 0.5849 \\
Arousal & Krippendorff's $\alpha$ (absolute distance) & 0.4951 \\
Dominance & Krippendorff's $\alpha$ (absolute distance) & 0.3138 \\
\bottomrule
\end{tabular}
\end{table}

\begin{table}[ht!]
\caption{Inter-annotator agreement and support for each emotion label.}
\label{tab:emotion_agreement_sorted}
\centering
\small
\begin{tabular}{lrrr}
\toprule
Emotion label & Fleiss' $\kappa$ & Krippendorff's $\alpha$  & Support (n)\\
\midrule
Neutral & 0.6506 & 0.6466 & 1219 \\
Desire & 0.6732 & 0.6726 & 450 \\
Gratitude & 0.9619 & 0.9619 & 407 \\
Annoyance & 0.3167 & 0.3114 & 190 \\
Disappointment & 0.3478 & 0.3461 & 112 \\
Anger & 0.5547 & 0.5547 & 103 \\
Joy & 0.4592 & 0.4587 & 98 \\
Confusion & 0.2792 & 0.2748 & 68 \\
Fear & 0.2733 & 0.2690 & 66 \\
Relief & 0.0739 & 0.0725 & 13 \\
Admiration & -0.0015 & -0.0012 & 4 \\
\bottomrule
\end{tabular}
\end{table}

Table~\ref{tab:iaa_general} reports these overall agreement scores for both multilabel emotion and dimensional annotations (valence, arousal, dominance). For emotion labels, we observe moderate agreement with a Krippendorff's $\alpha$ of 0.6090. Agreement for valence is similarly moderate ($\alpha$ = 0.5849), while arousal shows slightly lower reliability ($\alpha$ = 0.4951). Dominance yields the lowest agreement among the annotation tasks, with an $\alpha$ of 0.3138. These findings correspond to prior work establishing the difficulty of achieving high inter-annotator agreement in emotion annotation. For dimensional annotations, Wood et al.~\cite{Wood2018} and Labat et al.~\cite{Labat2024} report comparable Krippendorff's $\alpha$ scores, with valence showing the highest reliability and dominance the lowest.

Table~\ref{tab:emotion_agreement_sorted} in turn delves deeper into the inter-annotator agreement for individual emotion labels, along with the number of annotations per label (viz. support). Categories such as \textit{desire}, \textit{gratitude}, and \textit{neutral} exhibit relatively high inter-annotator agreement, with $\kappa$ and $\alpha$ above 0.64. In particular, \textit{gratitude} stands out with near-perfect agreement ($\kappa = 0.9619$, $\alpha = 0.9619$), probably linked to the explicit lexical cues commonly associated with the expression of this emotion. In contrast, emotions like \textit{admiration}, \textit{confusion}, \textit{fear}, and \textit{relief} show lower agreement, with $\kappa$ and $\alpha$ falling below 0.30, and even slightly negative in the case of \textit{admiration}. This suggests that these emotions are harder to annotate consistently, potentially due to their lower support or subtler linguistic manifestations. The remaining categories fall between these two extremes, with \textit{anger} and \textit{joy} showing moderate agreement ($\kappa$ of 0.5547 and 0.4592, respectively), while \textit{annoyance} and \textit{disappointment} yield lower, but still fair agreement ($\kappa$ of 0.3167 and 0.3478, respectively).

\subsubsection{Agreement Between Self-Reported and Third-Party Emotion Labels} 
\begin{figure}[h!]
\begin{center}
\includegraphics[width=0.53\textwidth]
{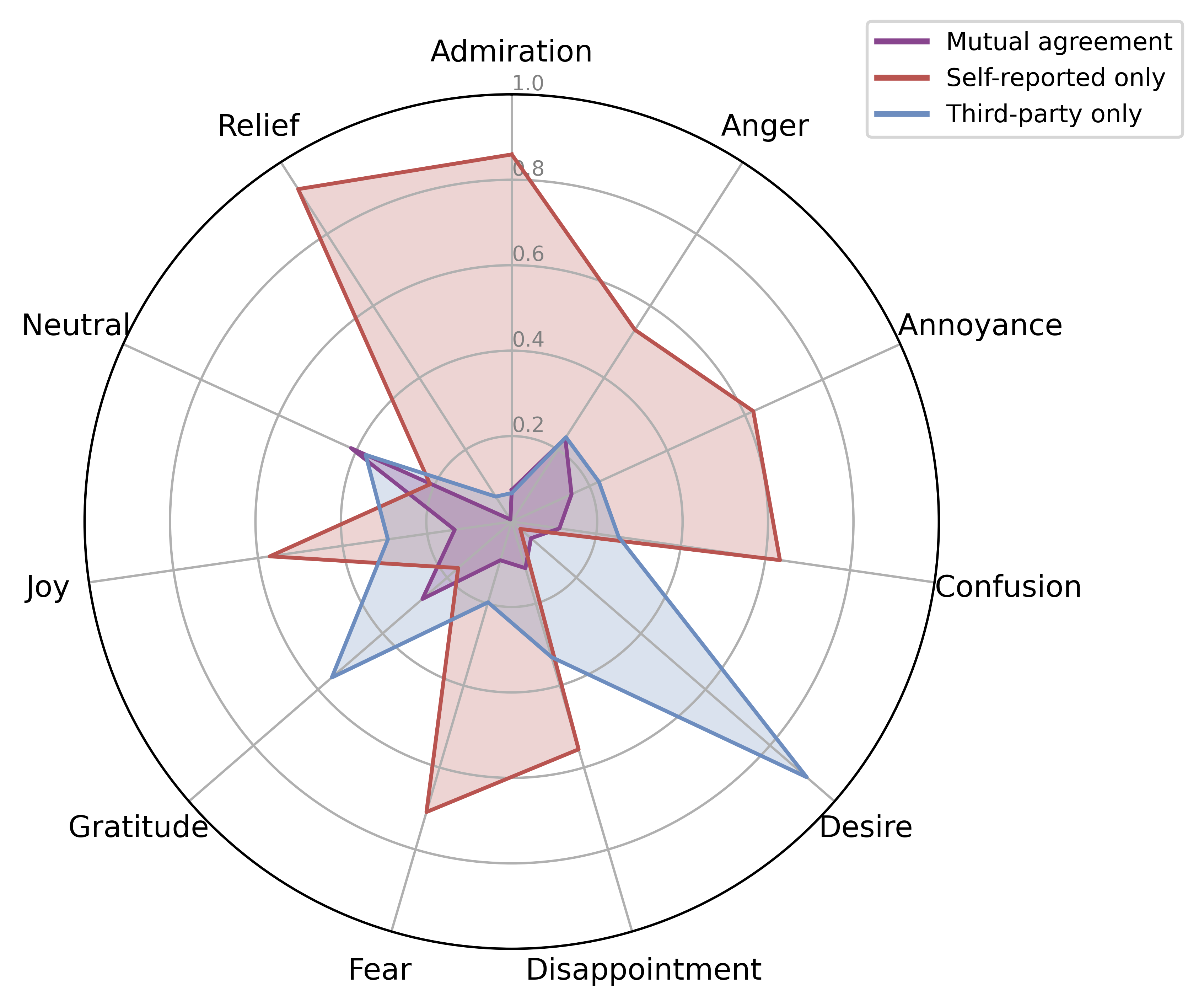}
\end{center}
\caption{Spider plot comparing emotion annotations between participants (self-reports) and third-party annotators. For each emotion category, the plot shows the average relative frequency of: mutual agreement (purple), only self-reported annotations (red), and only third-party annotations (blue). All frequencies are averaged across three independent third-party annotators.}
\label{fig:thirdparty_vs_selfreported_annot}
\Description{A radar (spider) chart comparing emotion labels from participants' self-reports and independent third-party annotators across multiple emotion categories. Each spoke represents a distinct emotion (e.g., admiration, fear, relief, desire, gratitude, neutral, anger, annoyance, joy). Three colored lines depict average relative frequencies: purple for mutual agreement between participant and annotator, red for emotions reported only by the participant, and blue for emotions labeled only by third-party annotators. Frequencies are computed per annotator and then averaged across three annotators. The plot highlights higher red values for emotions like admiration, fear, and relief, indicating these are often self-reported without external recognition. In contrast, desire and gratitude show higher blue values, suggesting annotators frequently label them without participant self-report, likely due to explicit lexical cues. A few categories, including neutral, gratitude, anger, annoyance, and joy, display comparatively higher purple agreement than others. Axes are normalized to relative frequency, and line thickness and markers are uniform; a legend maps colors to agreement conditions. The figure shows overall low agreement, with distinct gaps between the three lines varying by emotion.}
\end{figure}

Building on the previous analysis of third-party annotator agreement, we turn to a comparison between these external annotations and participants' self-reported emotions. 
Figure~\ref{fig:thirdparty_vs_selfreported_annot} visualizes how emotion labels align or diverge between the experiencers themselves and independent annotators. The spider plot visualizes the relative frequency of annotations across emotion categories, with each axis representing a distinct emotion. The plotted lines show the average proportion of instances for three scenarios: (1) mutual agreement between participant and annotator (purple); (2) emotions reported only by the participant (red); and (3) emotions labeled only by the third-party annotator (blue).

Proportions are computed per third-party annotator and then averaged across all three to ensure robustness. The plot reveals differences in emotion perception between the two groups. For example, \textit{admiration}, \textit{fear}, and \textit{relief} show high participant only reporting, with values above 0.70, suggesting that these emotions are more often self-perceived than externally recognized. In contrast, \textit{desire} and \textit{gratitude} are frequently only annotated by third-party observers (0.91 and 0.56 times, respectively), possibly due to the explicit lexical cues (e.g., \textit{thank you}, \textit{want}) associated with these emotions that annotators may have relied on during the annotation process. Overall, agreement between self-reported and third-party annotations is relatively low, which is further reflected by an average Krippendorff's $\alpha$ of 0.2065 (stdev=0.03) across pairwise comparisons between each annotator and the corresponding self-report. This low agreement underscores the subjective nature of emotional expression and perception, particularly in social interaction data. It also serves as a reminder of the inherent limitations of relying solely on observable cues to infer internal emotional states, especially for emotions that are subtly expressed. Nevertheless, a few categories such as \textit{neutral} (0.41), \textit{gratitude} (0.28), \textit{anger} (0.23), \textit{annoyance} (0.15), and \textit{joy} (0.13) show comparatively higher mutual agreement. This might be due to their clearer or more conventional linguistic markers that are both internally experienced and externally observable.

\begin{figure*}[ht!]
\begin{center}
\includegraphics[width=0.75\textwidth]
{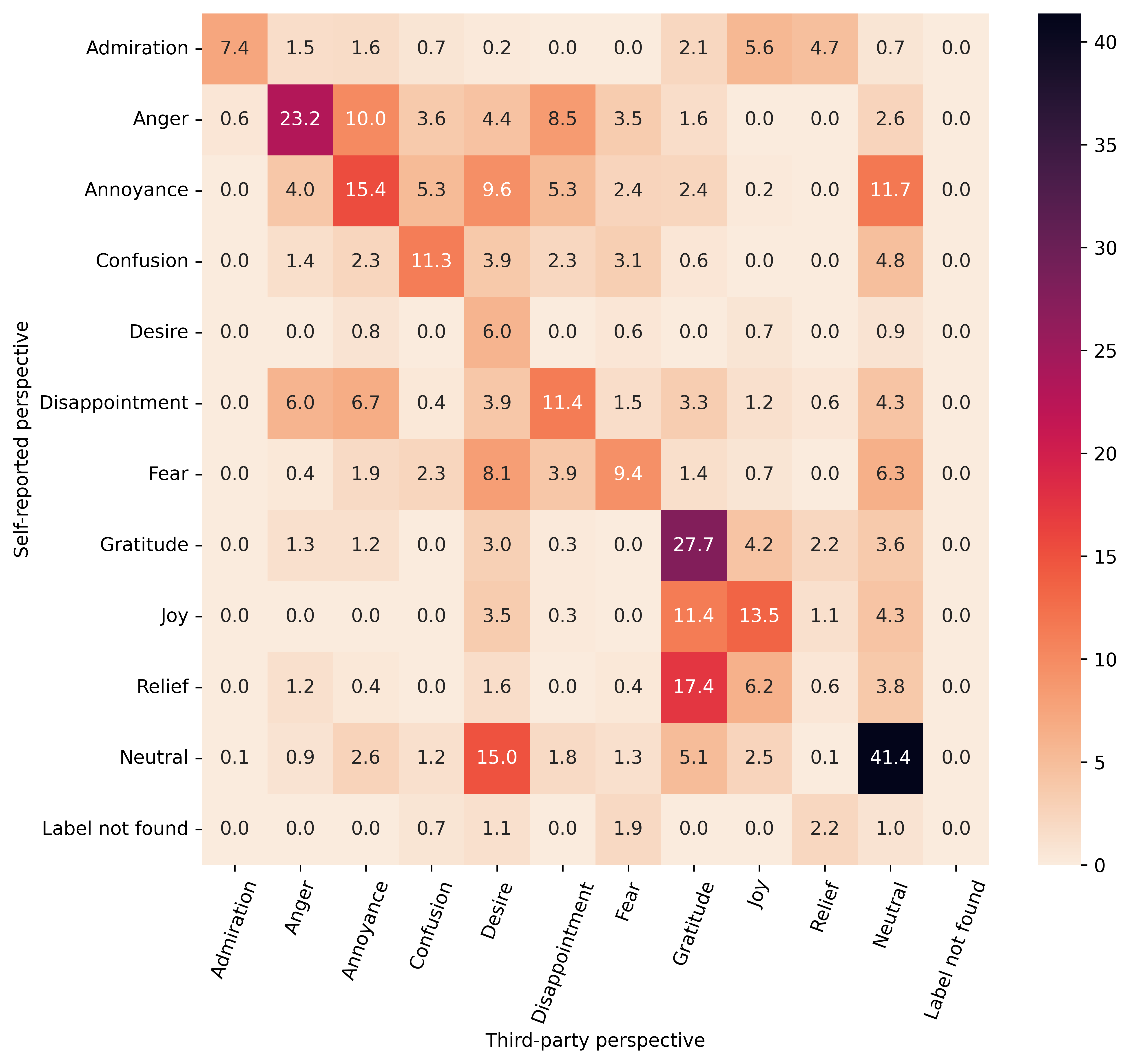}
\end{center}
\caption{Heatmap showing pairwise agreement between self-reported and third-party emotion annotations using Jaccard similarity (in \%). Each cell reflects average similarity between participant (y-axis) and annotator (x-axis) labels. Darker shades indicate stronger alignment.}
\label{fig:thirdparty_vs_selfreported_agreement}
\Description{A rectangular heatmap visualizing Jaccard similarity scores between participants' self-reported emotions (rows on the y-axis) and emotions labeled by any of three third-party annotators (columns on the x-axis). Each cell shows the average percentage overlap between the set of messages tagged with a given emotion by participants and the corresponding set tagged by external annotators, averaged across annotators. Darker cells indicate higher similarity, lighter cells indicate lower similarity. The diagonal cells reflect within-emotion agreement, with comparatively darker shades for neutral (41.4\%), gratitude (27.7\%), anger (23.2\%), and annoyance (15.4\%). Off-diagonal patterns indicate systematic confusions: participant-reported relief aligns with annotator gratitude (17.4\%) and joy (6.2\%); participant neutral aligns with annotator desire (15.0\%); participant annoyance aligns with annotator neutral (11.7\%). A category ``label not found'' appears on the participant side but is absent from annotator columns. Axis tick labels list discrete emotions (e.g., neutral, gratitude, anger, annoyance, joy, relief, desire, admiration, fear, etc.). A colorbar conveys similarity magnitude in Jaccard distance, with darker tones marking stronger overlap.}
\end{figure*}

While Figure~\ref{fig:thirdparty_vs_selfreported_annot} provides a high-level overview of annotation agreement across emotion categories, it does not capture \textit{which} specific emotions are being confused across perspectives. To explore these patterns of confusion in more detail, Figure~\ref{fig:thirdparty_vs_selfreported_agreement} presents a pairwise comparison of annotation overlap using the Jaccard similarity metric. This heatmap visualizes the degree of alignment between self-reported and third-party labels for each emotion pair, with similarity scores calculated between the sets of messages tagged with a given emotion by participants and those labeled by any of the three external annotators. Each cell in the heatmap represents the average similarity between internal and external annotations, aggregated across external annotators. Darker shades indicate stronger similarity, while lighter shades reflect divergence between emotion categories. The x-axis corresponds to emotions identified by external annotators, and the y-axis to those reported by participants themselves. As in Figure~\ref{fig:thirdparty_vs_selfreported_annot}, the heatmap reveals that agreement is highest for categories such as \textit{neutral} (41.4\%), \textit{gratitude} (27.7\%), \textit{anger} (23.2\%), and \textit{annoyance} (15.4\%). In contrast, several patterns of disagreement emerge. Emotions such as \textit{gratitude}, \textit{joy}, and \textit{relief} are often confused. For example, participants may report \textit{relief}, while external annotators are more likely to label the same instance as \textit{gratitude} (17.4\%) or \textit{joy} (6.2\%). Other mismatches include cases where participants report \textit{neutral} but annotators perceive \textit{desire} (15.0\%), or where \textit{annoyance} is self-reported but interpreted as \textit{neutral} (11.7\%) by third parties. Finally, the option \textit{label not found} was occasionally used in the self-reported perspective, but never by external annotators.

\subsection{Conversational Dynamics} \label{sec4:conv_dynamics}
This section explores how conversational dynamics unfold throughout the interaction. We begin by examining the various operator response techniques used by wizards and how these techniques relate semantically to one another. Next, we investigate how operator response strategies influence participant emotions, revealing associations between specific response strategies and their elicited emotional reactions. Finally, we analyze the progression of emotional states (captured in valence) over time, finding differences in emotional trajectories across conversations with different target valence categories.

\subsubsection{Operator Response Techniques}
A total of 12,848 wizard messages were collected, including fixed conversation starters (e.g., ``Hello, my name is Chatty1 \smiley \space How can I help you today?''), which were not annotated. After excluding these 2,147 fixed starters, one for each of the 179 conversations per 12 participants, 10,701 messages remained for annotation.\footnote{Due to a malfunction in the application, affecting participant ID 98, one conversation starter was not displayed, resulting in a slight discrepancy from the expected count.} The annotation process was integrated into the message writing workflow, as wizards classified their own messages before sending them. This approach allowed them to ensure that each message (or part thereof) was appropriately classified. Wizards could decide to divide a message into multiple parts based on its semantic content and the communicative intent or tone. For example, a message might begin with an empathetic acknowledgment and then transition into a factual explanation, warranting two separate parts with distinct labels. Each part was then labeled with a single overarching technique (i.e., \textit{affective}, \textit{objective}, \textit{suboptimal}, or \textit{alternative}) and one or more associated response strategies (see Table~\ref{tab:rs_strat_list} in Section~\ref{sec3:annot_rs}). Consequently, the annotation task followed a multilabel setup at the message level for both techniques and strategies. Table~\ref{tab:techniques_strategies_counts} provides an overview of the distribution of response strategies across the four main techniques in each message. Notably, the most frequently used strategies were \textit{explanation} (30.3\%) and \textit{request information} (28.4\%), both under the \textit{objective} technique. Among affective strategies, \textit{cheerfulness} (10.6\%) and \textit{empathy} (7.0\%) were most common. Suboptimal responses, while less frequent, still accounted for a significant portion (22.6\% combined), capturing areas where wizard responses deviated from ideal interaction patterns.

\begin{table}[ht!]
\caption{Distribution of response strategies grouped by technique across 10,701 annotated wizard messages. Each message may contain multiple parts, with each part annotated with a single technique and one or more associated response strategies.}
\label{tab:techniques_strategies_counts}
\centering
\small
\begin{tabular}{llrr}
\toprule
Technique & Response Strategy & Support (n) & Percentage (\%) \\
\midrule
\multirow{4}{*}{Affective} 
  & Apology & 332 & 3.1 \\
  & Cheerfulness & 1,138 & 10.6 \\
  & Empathy & 749 & 7.0 \\
  & Gratitude & 276 & 2.6 \\
\cmidrule(r){1-1}\cmidrule(r){2-2}\cmidrule(r){3-3}\cmidrule{4-4}
  \multirow{5}{*}{Objective} 
  & Explanation & 3,242 & 30.3 \\
  & Help offline & 409 & 3.8 \\
  & Help online & 1,293 & 12.1 \\
  & Request action & 649 & 6.1 \\
  & Request information & 3,037 & 28.4 \\
\cmidrule(r){1-1}\cmidrule(r){2-2}\cmidrule(r){3-3}\cmidrule{4-4}
\multirow{4}{*}{Suboptimal} 
  & Inappropriate response & 723 & 6.8 \\
  & Irony & 274 & 2.6 \\
  & Miscomprehension & 397 & 3.7 \\
  & Non-collaborative response & 1,019 & 9.5 \\
\cmidrule(r){1-1}\cmidrule(r){2-2}\cmidrule(r){3-3}\cmidrule{4-4}
\multirow{1}{*}{Alternative} 
  & Other & 7 & 0.1 \\
\bottomrule
\end{tabular}
\end{table}

\begin{figure*}[h!]
\begin{center}
\includegraphics[width=\textwidth]
{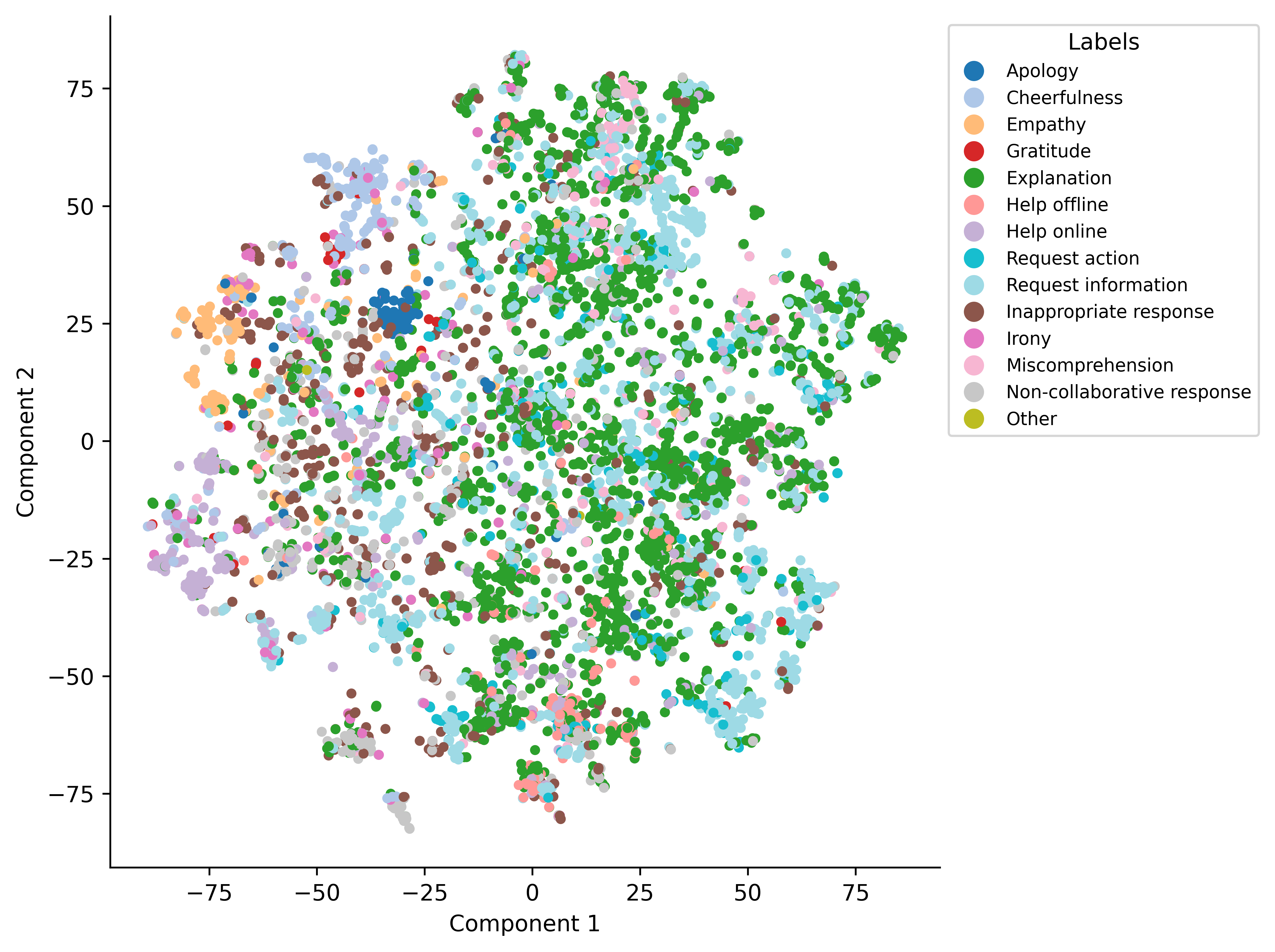}
\end{center}
\caption{Two-dimensional visualization of sentence embeddings for response strategies. Each point represents a part, color-coded according to its labeled response type.}
\label{fig:cluster_embedding_rs}
\Description{A two-dimensional scatterplot created with t-SNE shows sentence embeddings of 8,894 unique Dutch utterances, each point representing a single-labeled segment mapped from a 256-dimensional embedding space to two dimensions. Points are color-coded by response strategy, with visually uneven class densities indicating imbalanced category frequencies; dense fields of points correspond to dominant strategies such as explanation and request information. Several affect-related strategies—apology, empathy, cheerfulness, and gratitude—appear as relatively compact, well-separated clusters, suggesting consistent phrasing and shared affective features. Other strategies form more diffuse or overlapping regions, indicating broader linguistic variability or semantic proximity with the conversation context. No duplicate texts or multi-labeled segments are included, ensuring each point has one color. The layout includes clearly legible axis tick marks without semantic meaning (t-SNE axes), a legend mapping colors to strategy names, uniform point size and opacity to reduce overplotting, and even spacing without grid lines. The visualization emphasizes overall structure: tight, small clusters for affective strategies; larger, heterogeneous clouds for high-frequency, information-oriented strategies; and areas of partial adjacency where semantically related strategies lie near each other.}
\end{figure*}

To better understand the relationships between different response strategies, we visualized their corresponding sentence embeddings for individual parts in a two-dimensional space, as shown in Figure~\ref{fig:cluster_embedding_rs}. Each point is color-coded according to its associated response strategy. To ensure clarity in the visualization, we included only utterance segments labeled with a single response strategy -- excluding multi-labeled instances to maintain consistent color coding. Additionally, we retained only unique utterances, removing duplicates to facilitate dimensionality reduction via t-SNE. This filtering process reduced the original set of 14,081 annotated segments to 8,894. Dutch utterances were embedded using a sentence embedding model pre-trained on RobBERT~\cite{Delobelle_Remy_2024}, an open-source Dutch language model.\footnote{The sentence embedding model trained on RobBERT is publicly available at \url{https://huggingface.co/FremyCompany/stsb_ossts_roberta-large-nl-oscar23}.} We opted for a sentence embedder rather than a regular language model, as sentence embeddings are specifically optimized for semantic similarity tasks such as clustering and search. In other words, these models are trained to maximize semantic alignment between related texts~\cite{reimers2019}. The sentence embedding model mapped each utterance to a dense 256-dimensional vector space, which was subsequently reduced to two dimensions using the t-SNE algorithm~\cite{vandermaarten2008}.

The scatterplot in Figure~\ref{fig:cluster_embedding_rs} offers insight into the semantic landscape of response strategies, capturing nuanced communicative intents. A first, prominent observation is the imbalance in class distribution. Although the figure only includes points with unique labels and texts, the dominance of categories such as \textit{explanation} and \textit{request information} reflects the trends reported in Table~\ref{tab:techniques_strategies_counts}. The affective techniques (viz. \textit{apology}, \textit{empathy}, \textit{cheerfulness}, and \textit{gratitude}) form relatively compact and distinct clusters. This suggests that the sentence embedding model effectively captures consistent linguistic patterns across utterances with similar communicative intents. Their tight grouping likely stems from both limited lexical variation (e.g., few ways to phrase an apology) and their topic-independent nature. The close proximity of these strategies in the embedding space further reflects shared affective qualities.

In contrast, the remaining techniques are more diffusely distributed across the embedding space, indicating broader applicability and stronger dependence on conversational context. This diffuse pattern corresponds to a semantic continuum where embeddings encode not only the communicative function, but also topical overlap. For example, points in the upper y-axis and positive x-axis regions often relate to travel and flights, while those clustered around coordinates (75, 25) predominantly concern telecommunication-related topics. Strategies such as \textit{explanation}, \textit{request action}, \textit{request information}, and \textit{miscomprehension} appear to be strongly topic-associated with their linguistic realization being influenced by domain-specific content. In contrast, the \textit{help online} strategy (e.g., ``I'll try to help you the best I can'') forms a relatively tight cluster adjacent to the affective techniques, probably due to its supportive tone and general-purpose phrasing. Even though the suboptimal strategies \textit{inappropriate response} and \textit{non-collaborative response} are scattered throughout the embedding space, they tend to occur more frequently near these affective and \textit{help online} regions. For instance, non-collaborative utterances such as ``I can't help you with that'' are positioned close to the \textit{help online} cluster.

\subsubsection{Operator Response--Emotion Interactions}
\begin{figure*}[h!]
\begin{center}
\includegraphics[width=0.77\textwidth]
{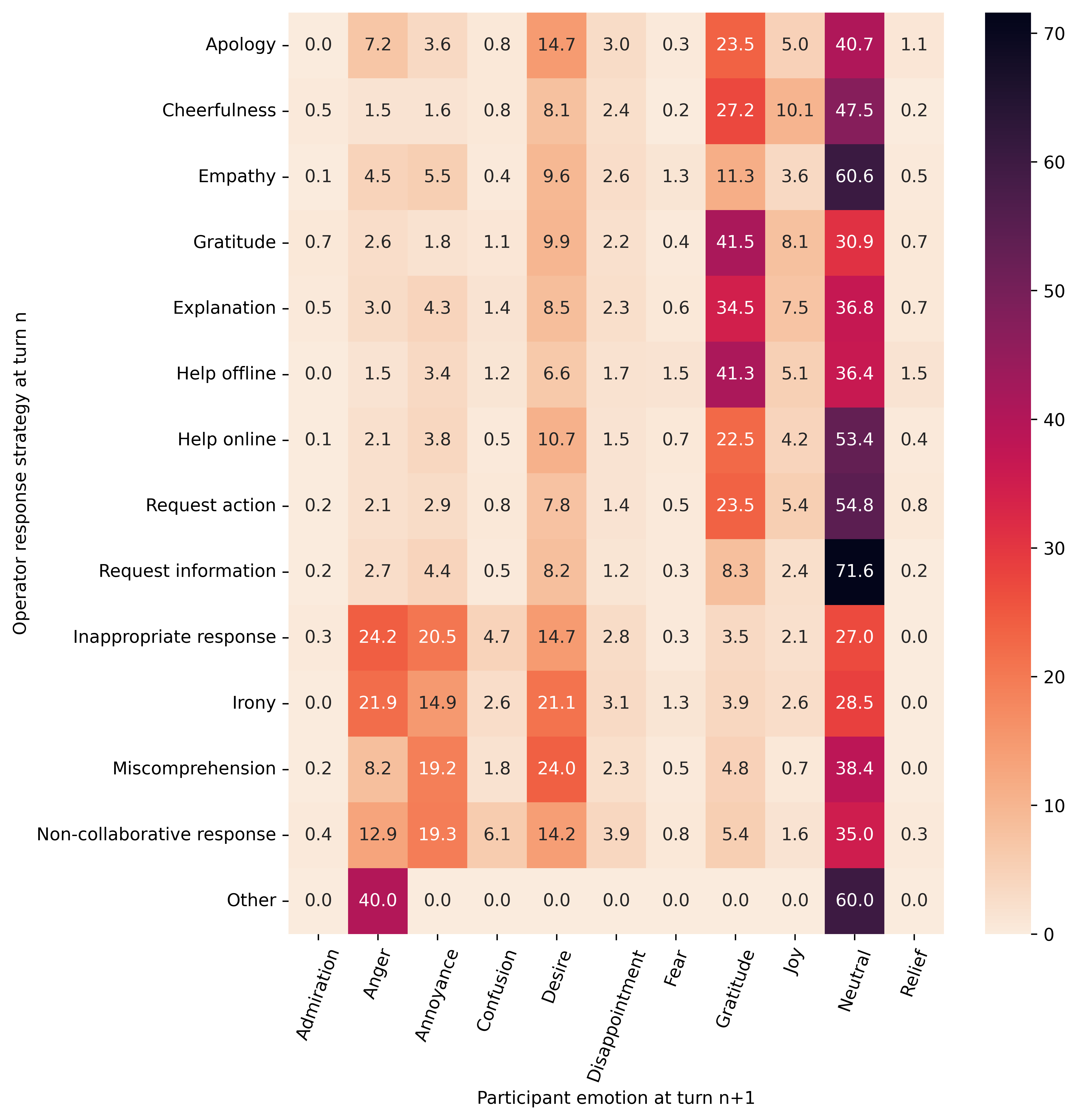}
\end{center}
\caption{Heatmap of normalized emotion distributions (in \%) across response strategies. Each cell reflects the relative prevalence of a specific emotion within a given strategy. Darker shades indicate higher relative frequency.}
\label{fig:heatmap_rs_emo}
\Description{A rectangular heatmap showing normalized percentages of participant emotions that occur after different operator response strategies. Rows (or columns) list response strategies such as request information, explanation, help online, help offline, request action, empathy, apology, gratitude, cheerfulness, other, and several suboptimal strategies including inappropriate response, irony, miscomprehension, and non-collaborative response. The opposite axis lists emotions including neutral, gratitude, joy, desire, anger, annoyance, disappointment, confusion, and others. Cell shading darkens with higher relative frequency within each strategy. Most strategies show a dark cell for neutral, with particularly strong intensity after request information, empathy, other, request action, and help online. Gratitude is most intense for the strategies gratitude and help offline, and appears prominently -- though slightly less than neutral -— after explanation. Cheerfulness as a strategy is followed by increases in gratitude and joy. Apology shows a mixed pattern with notable concentrations of gratitude, desire, anger, and annoyance. Empathy exhibits modest gratitude and joy alongside desire, anger, and annoyance. Suboptimal strategies present distinct darker bands in negative or need-indicative emotions: inappropriate response aligns strongly with anger and annoyance; miscomprehension with desire and annoyance; irony and non-collaborative response with higher anger, annoyance, disappointment. A single colorbar indicates relative percentages, with darker shades indicating higher within-strategy prevalence for the corresponding emotion.}
\end{figure*}

As operator response strategies were employed to steer participant emotions, this section explores how specific strategies tend to evoke different emotional reactions. To this end, we extracted all wizard messages from the dataset that were followed by a participant reply. As dialogue structure does not follow a strict alternation between wizard and participant messages (viz. either party may contribute multiple messages before receiving a response), we grouped uninterrupted consecutive messages from the same interlocutor into a single turn. This grouping was applied to both wizard and participant messages to ensure consistency. We then paired each wizard turn (containing response strategies) with the subsequent participant turn (containing emotions) to create parallel lists: one of wizard strategies and one of participant emotions. For each strategy, we counted the emotion labels in the paired participant reply and normalized these counts by the total number of emotion labels observed for that strategy. Figure~\ref{fig:heatmap_rs_emo} presents the distribution of participant emotions following each operator response strategy.

From this figure, we learn that \textit{neutrality} mostly dominates as the default state expressed in participant turns as response to the majority of response strategies, peaking particularly after \textit{request information} (71.6\%), \textit{empathy} (60.6\%), \textit{other} (60\%), \textit{request action} (54.8\%) and \textit{help online} (53.4\%), suggesting that these strategies either (partially) address participant concerns or fail to elicit strong affective responses. The strategies \textit{gratitude} and \textit{help offline} are the only ones in which the emotion \textit{gratitude} surpasses \textit{neutrality} in frequency, while the strategy \textit{explanation} also elicits substantial levels of \textit{gratitude} (34.5\%), albeit slightly less than \textit{neutrality} (36.8\%). Gratitude expressed in response to \textit{help offline} and \textit{explanation} may reflect participants' recognition of effortful or informative support, suggesting that these strategies are perceived as more considerate or helpful in addressing user needs. In the case of \textit{help offline} specifically, the expression of gratitude might also signal the closure of the interaction, where participants thank the wizard and are referred to a human operator for further assistance. Moreover, when the operator expresses gratitude, it appears to foster positive reciprocity in participants.

The same principle applies to \textit{cheerfulness} which is frequently preceded by \textit{gratitude} (27.2\%) and \textit{joy} (10.1\%), but less so to the other affective techniques \textit{apology} and \textit{empathy}. While \textit{apology} elicits a considerable amount of \textit{gratitude} (23.5\%), it is also linked to higher levels of \textit{desire} (14.7\%), \textit{anger} (7.2\%), and \textit{annoyance} (3.6\%), suggesting that apologies may not consistently resolve participant concerns or foster positive affect. Similarly, \textit{empathy} results in a relatively low expression of \textit{gratitude} (11.3\%) and \textit{joy} (3.6\%), with \textit{desire} (9.6\%), \textit{anger} (4.5\%), and \textit{annoyance} (5.5\%) remaining present. These findings imply that while \textit{apology} and \textit{empathy} are affective in nature, they may not be perceived as sufficiently impactful or emotionally resonant within customer service interactions where participants believe they are engaging with an autonomous system, as in the WOZ setup. Finally, a clear distinction emerges between directly supportive strategies and suboptimal techniques, namely \textit{inappropriate response}, \textit{irony}, \textit{miscomprehension}, and \textit{non-collaborative response}, which yield distinctly different emotional profiles. Suboptimal techniques are associated with a marked increase in negative affect, including \textit{anger}, \textit{annoyance}, \textit{confusion}, \textit{desire}, and \textit{disappointment}. Collectively, these emotions account for over 55\% of participant responses to suboptimal strategies, with some individual response--emotion pairs exceeding 20\% (e.g., \textit{inappropriate response} followed by \textit{anger} in 24.2\% of cases; \textit{miscomprehension} followed by \textit{desire} in 24.0\%). Notably, beyond the low-valence emotions \textit{anger}, \textit{annoyance}, and \textit{disappointment}, \textit{desire} (14.2--24.0\%) and \textit{confusion} (1.8--6.1\%) also show a moderate surge after suboptimal replies. This pattern suggests unresolved needs or dissatisfaction, underscoring the importance of clear and emotionally attuned engagement in customer service.

\subsubsection{Progression of Emotional States}
To examine how emotional states develop across interactions with different target valence conditions, we segmented all conversations into three sets according to the target valence for wizards: positive, neutral, negative. For reliable temporal analysis, we set a minimum inclusion criterion of four participant messages per conversation, ensuring that each temporal segment would contain at least one message and allowing for a meaningful breakdown over time. Following filtering, 69.3\% (n=496) of positive-target conversations, 60.9\% (n=437) of neutral-target conversations, and 68.7\% (n=491) of negative-target conversations were retained for analysis. Each conversation was split into four chronological bins (early, mid-early, mid-late, and late) by allocating conversation messages as evenly as possible across phases, ensuring bin sizes differ by at most one message. We then calculated the average proportion of valence types per bin, and weighted the contribution of each conversation equally, such that lengthy conversations do not disproportionately impact the results. The approach enables meaningful comparison of conversational dynamics across different stages regardless of original conversation lengths. The results from this approach for valence are illustrated in Figure~\ref{fig:val_progression}. Similar plots for arousal and dominance are available in Appendix~\ref{app:ar-dom_progression}.

\begin{figure*}[h!]
\begin{center}
\includegraphics[width=0.8\textwidth]
{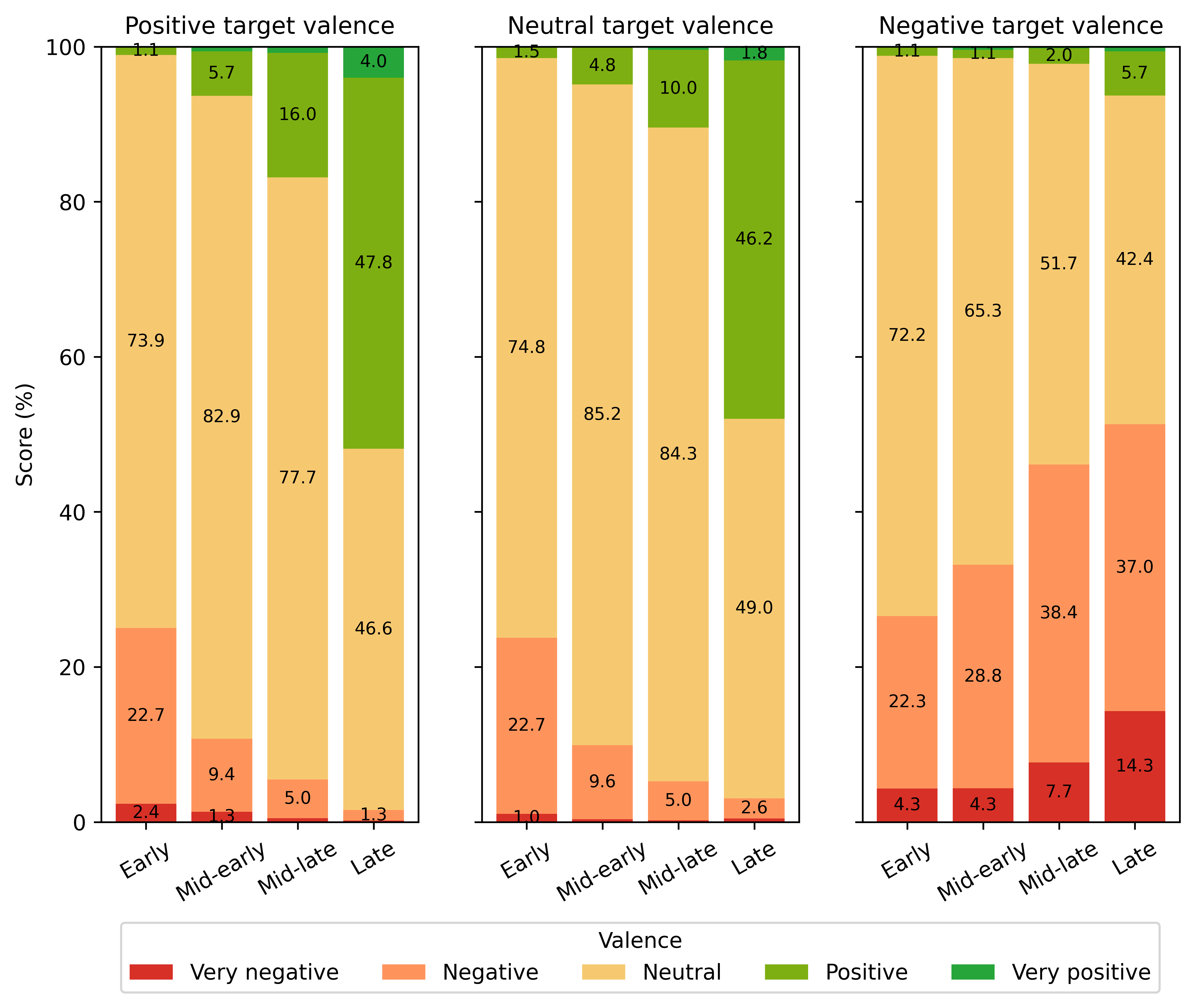}
\end{center}
\caption{Temporal progression of expressed valence in interactions, grouped by target valence (positive, neutral, negative). Stacked bar charts show valence distribution from very negative (deep red) to very positive (dark green) across four temporal bins. Percentage labels mark segments above 1\%. The four chronological bins depict valence as it evolves over the course of the interaction.}
\label{fig:val_progression}
\Description{Three sets of grouped stacked bar charts illustrating how participant valence changes over time within conversations assigned to positive, neutral, or negative target conditions. For each target condition, four adjacent bars represent chronological bins (early, mid-early, mid-late, late), with each bar segmented into ordered valence slices from very negative (deep red) through neutral to very positive (dark green). Segment percentages are printed on slices exceeding 1\%. The positive-target panel shows a gradual increase in positive-toned segments toward the late phase; the neutral-target panel shows a similar trend, although the share of neutral for each timestamp is consistently higher; the negative-target panel shows growing negative proportions in later bins. The three panels are visually comparable, using the same color scale and bin structure, enabling side-by-side inspection of trajectories. Sample sizes across bins in the same panel are controlled by weighting the contribution of each conversation equally and evenly partitioning messages into bins that differ by at most one message. The layout emphasizes temporal evolution rather than raw counts, with a red–green valence scale and explicit labels for the four temporal bins per target condition.}
\end{figure*}

Across the three scenarios, we observe gradual and stable shifts in valence, reflecting our goal of the experiment to direct conversations toward specific target outcomes. This steering effect becomes gradually pronounced as conversations progress: 51.8\% of those with a positive target end positively, 49.0\% of those with a neutral target conclude with a neutral tone, and 51.3\% of those targeting negativity finish in a negative emotional state. The distributions of valence for positive and neutral targets are highly similar, although positive-target conversations have a slightly larger increase in positive outcomes. In contrast, conversations with a negative target valence display a trend toward heightened negativity rather than a reduction thereof. During the data collection, wizards mentioned that distinctly steering conversations between positive and neutral emotional outcomes was challenging. Interestingly, for our customer interactions, Figure~\ref{fig:val_progression} shows that there is a minimal difference in results between a highly engaged approach, seeking to uplift the customer's emotional state, and a less engaged one. Lastly, while our experiment, which focused on emotion shifts across entire conversations, was mostly successful, future research could explore tracking and influencing emotional shifts within smaller conversational segments by introducing multiple, evolving target valence points throughout the interaction.

\section{Automatic Emotion Inference in Social Interaction}\label{sec5:ML}
Having described the design and properties of the \corpus~corpus, we now turn to automatic emotion inference. Emotion detection aims to identify the current emotional state of a speaker, whereas emotion inference goes further by anticipating how that state may evolve as the interaction unfolds. Our goal is to evaluate how well modern machine learning systems infer emotions for unseen, future turns in customer service dialogues. Section~\ref{sec5:ML_lit} reviews related work on emotion modeling with context. Section~\ref{sec5:ML_task} formally introduces the classification task of emotion inference; while Section~\ref{sec5:ML_setup} details our experimental setup. Finally, Section~\ref{sec5:ML_results} presents the results of our experiments.

\subsection{Related Work on Emotion Modeling with Context}\label{sec5:ML_lit}
Traditionally, automatic emotion detection has focused on classifying the emotional state of single text units drawn from sources such as blogs, news articles, or social media~\cite{bostan-klinger-2018-analysis}. These approaches typically operate in isolation, without accounting for additional conversational or situational information that might shape the interpretation of emotion. Although single-instance classification can be effective, it overlooks contextual cues that are valuable in interactive settings such as dialogues, where emotions unfold across turns rather than in isolation. In conversation, an utterance often gains emotional meaning when considered in relation to previous exchanges, the speaker's trajectory, or the dialogue partner's responses. Consequently, research in this field has increasingly moved toward contextualized models for emotion recognition that incorporate the surrounding conversational context and its evolving dynamics~\cite{Poria2019}.

A growing body of research has explored different algorithms and methodologies for incorporating context into emotion recognition in conversation. An important contribution in this direction was the SemEval-2019 Task 3, which introduced EmoContext~\cite{Chatterjee2019}, a benchmark that explicitly required participants to classify the emotion of the current utterance while considering the two preceding utterances for contextualization. Over the years, approaches to modeling emotions with surrounding context have progressed from recurrent neural architectures that encode multi-utterance dependencies and speaker states~\cite{Majumder2019,Li2022}, to graph-based models that propagate signals over utterance-speaker interaction graphs~\cite{Ghosal2019,Shen2021}, and to transformer-based approaches that capture long-range dependencies and discourse structure~\cite{wemmer-etal-2024-emoprogress}. These frameworks are often combined in hybrids (e.g., graph-enhanced transformers, hierarchical transformers, and RNN-graph or RNN-transformer pipelines), with some models further leveraging external knowledge bases to enrich contextual cues and disambiguate emotion causes~\cite{Zhao2022p628,Li_Zhu_Mao_Cambria_2023}. Even more recently, instruction or prompt-driven paradigms have gained traction~\cite{Zhang2025}. Although context and external knowledge often aids emotion detection, researchers found that some utterances are clear on their own, addressing the need to identify when context is required for disambiguation~\cite{tu-etal-2023-context,wemmer-etal-2024-emoprogress}.

In contrast to the majority of work on emotion recognition in observed utterances, emotion inference aims to predict the emotional state of an interlocutor's upcoming turn using only prior conversational context. By shifting the focus from recognition to anticipation of emotional dynamics, this task is especially relevant for real-time applications where systems must respond adaptively and prevent harmful interactions before they escalate. Early research in this area focused on predicting the emotions of addressees in online dialogue using n-grams and emotion features~\cite{Hasegawa2013-predicting}. Subsequent work on sentiment or valence forecasting introduced recurrent neural architectures to estimate the upcoming utterance's polarity~\cite{Bothe2017} and by jointly simulating the next utterance from prior context while modeling hierarchical contextual influence~\cite{Wang-2020-sentiment}. More recent inference models conceptualize emotion prediction as 
leveraging dialogue context and role information: the IDSEC architecture links conversational history with latent emotional states for next-turn prediction, capturing emotion persistence and contagion~\cite{li2020-interactive}; an addressee-aware ensemble incorporates speaker-listener roles and aggregates multiple predictors for multi-turn inference~\cite{li-etal-2021-emo-inf}; and external commonsense knowledge enables reasoning beyond surface cues~\cite{li2021-enhancing-emo-inf}.

\subsection{Our Problem Definition of Emotion Inference}\label{sec5:ML_task}
Building on the contextualized approaches above, we shift from recognizing emotions in observed utterances to inferring the customer's next-turn emotion from prior dialogue. Our aim is to benchmark \corpus~for emotion inference and clarify how context can be leveraged in our setup. We isolate two complementary sources of context, viz.~the interlocutor's previous turn and the customer's last turn, and assess their individual and fused contributions. We now formalize the task and the model variants for controlled comparison. A dialogue is represented as a sequence of textual \textit{turns}, where each turn consists of one or more consecutive messages produced by the same interlocutor until the other party responds. We distinguish between customer turns (produced by our participants) and operator turns (produced by wizards). Emotion is annotated only for customer turns; although the annotation scheme includes categorical emotions, valence, arousal, and dominance, our experiments use only the categorical emotion and ordinal valence annotations. For emotion inference, the customer's current turn $t_i$ is not observed. Instead, the system must infer the future emotion $y_i$ based solely on prior conversational context. This formulation differs from standard emotion classification tasks, which rely directly on the text of the instance being labeled.

Formally, a dialogue is defined as a sequence of turns 
\[
\mathcal{D} = \{ t_1, t_2, \dots, t_N \},
\] 
where $N$ denotes the total number of turns. Let $\mathcal{E}$ denote the set of emotion categories and $\mathcal{V}$ the set of ordinal valence levels. We write the task-specific label space as $\Omega \in \{\mathcal{E}, \mathcal{V}\}$ and the target variable as $y_i \in \Omega$, where $\Omega$ is selected per run to match the chosen label column. For each non-initial customer turn $t_i$, we aim to predict $y_i$ before $t_i$ is observed. We leverage two prior textual contexts:

\[
\begin{aligned}
C_i^{\text{opp}} &= t_{i-1} && \text{(previous turn by the other interlocutor)}, \\
C_i^{\text{self}} &= t_{i-2} && \text{(last turn by the same interlocutor as $t_i$)}.
\end{aligned}
\]
We apply this procedure to all customer turns except the first, since the required prior contexts are not fully available. We instantiate two base predictors and combine them via weighted fusion:
\begin{itemize}
\item Model 1 (other-to-future) uses $C_{i}^{\text{opp}}$. The model captures how the other interlocutor's immediately preceding turn shapes the customer's forthcoming affect:
\[
\begin{aligned}
p_{i}^{(1)}(y) &= f_1(t_{i-1})
&& \text{predicts } p(y_i = y \mid t_{i-1}), \quad y \in \Omega.
\end{aligned}
\]

\item Model 2 has two variants using $C_i^{\text{self}}$, namely 2a (self-current) and 2b (self-to-future). Model 2a performs emotion detection on the prior customer turn $t_{i-2}$ to identify the historical emotional state $y_{i-2}$, which serves as a persistence prior for predicting $y_i$. In contrast, Model 2b projects the customer's prior affect directly into their next turn:
\[
\begin{aligned}
p^{(2a)}_i(y) &= f_{2a}(t_{i-2})
&& \text{predicts } p(y_{i-2} = y \mid t_{i-2}), && y \in \Omega,\\
p^{(2b)}_i(y) &= f_{2b}(t_{i-2})
&& \text{predicts } p(y_i = y \mid t_{i-2}), && y \in \Omega.
\end{aligned}
\]
\end{itemize}
At inference time, we combine Model 1 with either Model 2a or 2b via late fusion. We combine the models' logits via a weighted sum and apply $\argmax_y$ on this sum to obtain the combined decision scores over $y$. The variable $k$ indexes the Model 2 variant (2a or 2b), while the fusion weight $\lambda$ is tuned on a held-out development set:
\begingroup
\setlength{\abovedisplayskip}{4pt}
\setlength{\belowdisplayskip}{4pt}
\begin{align*}
\hat{p}_i(y) &= \lambda p_{i}^{(1)}(y) + (1 - \lambda) p^{(2k)}_i(y), \quad k \in \{a, b\}, \quad \lambda \in \{0.1, 0.2, ..., 0.9\}
\end{align*}
\begin{align*}
\hat{y}_i = \argmax_{y \in \Omega} \text{}\hat{p}_i(y).
\end{align*}
\endgroup

As a reference ceiling, we also define a turn-level detector that operates on the (unobserved) current customer turn $t_i$ to perform emotion detection. The system $p_i^{(0)}$ constitutes an upper-bound for emotion inference because it has direct access to $t_i$:

\[
\begin{aligned}
p_{i}^{(0)}(y) &= f_0(t_{i})
&& \text{predicts } p(y_i = y \mid t_{i}), \quad y \in \Omega.
\end{aligned}
\]

\subsection{Experimental Setup}\label{sec5:ML_setup}
This section describes our dataset preprocessing and participant-level splits, the experimental organization and evaluation protocols, and implementation details for the different models. 

\noindent \paragraph{Preprocessing and Data Splits.} \corpus~comprises 23,706 utterances grouped into speaker-contiguous turns, yielding 20,128 single-speaker turns; labels were correspondingly merged per turn to align with the grouping. For valence, multiple ordinal scores within a turn were collapsed to a single integer via a heuristic that keeps the most extreme polarity label and resolves symmetric ties toward negativity: 1 > 5 > 2 > 4 > 3. Emotion annotation was originally multilabel; after turn grouping, only 10.5\% of customer turns had more than one non-neutral tag, motivating a reduction to multiclass using an inverse-frequency precedence, ensuring rarer emotions persist in the final targets: \textit{fear} > \textit{confusion} > \textit{joy} > \textit{disappointment} > \textit{anger} > \textit{annoyance} > \textit{gratitude} > \textit{desire} > \textit{neutral}. Simultaneously, the rare labels \textit{admiration} and \textit{relief} were mapped to \textit{joy} (32 instances each). We then construct context windows as trios $[t_{i-2}, t_{i-1}, t_i]$ where $t_{i-2}$ and $t_i$ have emotion annotations, producing 7,349 instances for sequence-conditioned modeling. For evaluation, we use 5-fold cross-validation at the participant level: all turns from a given participant are assigned to a single split (train, dev, or test) within each fold to prevent leakage across splits. Partitioning is performed over participants rather than individual turns, so split sizes can vary slightly because the 0.65/0.15/0.20 quotas are applied to participant counts, not instance counts. This results in five fixed train--dev--test configurations that cover the full dataset across folds. Details on per-fold instance totals are reported in Table~\ref{tab:data_splits_exp} of Appendix~\ref{app:exp}.

\paragraph{Experimental Design and Evaluation Metrics.} Experiments are organized into three groups to quantify component contributions while avoiding redundancy between baselines and ablations in the fused system. Group 1 (baselines): Single-model references include an upper-bound emotion detection model and the standalone predictors $p_{i}^{(1)}(y)$, $p^{(2a)}_i(y)$, and $p^{(2b)}_i(y)$. These baselines are evaluated on the held-out test splits (after tuning on dev sets) and serve as single-predictor references in the fusion setup. Group 2 (fusion): We combine $p_{i}^{(1)}(y)$ with either $p^{(2a)}_i(y)$ or $p^{(2b)}_i(y)$ via late fusion. To avoid duplication with the baselines of Group 1, ablations are reported as deltas relative to the full fusion and explicitly mark $\lambda=0$ (or branch deactivation), thus isolating each component's contribution within the composite model. Group 3 (zero-shot LLM): Finally, the best-performing fusion model is compared to zero-shot LLM prompting, which follows task instructions without in-context examples, to assess encoder fine-tuning against the zero-shot capabilities of LLMs.

To evaluate the results, we report accuracy, F1 macro, and F1 weighted on the held-out test split of each fold. Accuracy summarizes overall correctness but can be misleading under class imbalance, which applies to \corpus. F1 macro computes the unweighted mean of per-class F1, emphasizing performance on minority and difficult classes irrespective of support. F1 weighted averages per-class F1 using class support, producing a dataset-level summary comparable to accuracy while also reflecting precision-recall trade-offs. Final scores are computed per fold and reported as the mean and standard deviation across the 5 folds.

\paragraph{Implementation Details.} Both emotion detection/prediction over $\mathcal{E}$ and valence detection/prediction over $\mathcal{V}$ are treated as single-label multiclass classification on $\Omega$, addressed by fine-tuning pretrained encoder-only transformer models. For Dutch, the encoder is RobBERT-2023-base\footnote{\url{https://huggingface.co/DTAI-KULeuven/robbert-2023-dutch-base}} (the latest RoBERTa-style Dutch model), and for English, the encoder is ModernBERT-base\footnote{\url{https://huggingface.co/answerdotai/ModernBERT-base}} (a modernized BERT model); both models are loaded via the Transformers library. Inputs are tokenized with padding and truncation to maximum of 512 tokens. Training follows the fixed 5-fold participant-level protocol described above, with (hyper)parameters tuned on the development split and final metrics reported on each fold's held-out test split. Fine-tuning uses the optimizer AdamW with learning rate $3\mathrm{e}{-5}$, weight decay 0.01, linear warmup of 500 steps, batch size of 16, and up to 10 epochs with early stopping (patience 3) based on development accuracy. All experiments ran on a single NVIDIA Tesla V100-SXM2-16GB GPU. For late fusion, logits from the two task-specific heads are combined by a weighted sum followed by $\argmax$; the fusion weight $\lambda$ is tuned on the development split. For zero-shot evaluation, OpenAI's GPT-5 nano is prompted with a single instruction-only prompt on the same textual context as the weighted fusion model. Details on the prompt formulation are in Table~\ref{tab:prompt-spec-en-nl} of Appendix~\ref{app:exp}.

\subsection{Results of Experiments on \corpus}\label{sec5:ML_results}
We now present the findings of our machine learning experiments. The baseline results in Table~\ref{tab:exp_baselines} demonstrate clear performance differences across contextual approaches for emotion and valence prediction. As expected, the upper-bound model $p_{i}^{(0)}$ using the current turn for emotion detection achieves the strongest performance (accuracy: 74.4\% emotions, 76.2\% valence for NL), establishing a ceiling for contextual prediction methods. Similarly, the persistence-based approach $p_{i}^{(2a)}$ shows strong emotion detection capabilities on prior turns (71.8\% accuracy for emotions, 80.1\% for valence in NL), indicating that historical emotional states can be reliably identified from text. In contrast, the cross-turn prediction model $p_{i}^{(1)}$ shows somewhat lower performance (53.2\% emotion accuracy for NL), with particularly low F1-macro scores (0.167), indicating the difficulty of predicting customer emotions from wizard responses alone. The future projection variant $p_{i}^{(2b)}$ performs worst across all metrics (48.5\% emotion accuracy for NL), demonstrating that direct temporal projection of prior emotional states is not always effective for future turn prediction. Valence classification consistently outperforms emotion classification by several percentage points across all models, while cross-linguistic differences remain minimal (<1\% accuracy variance), suggesting robust predictive patterns across both Dutch and English interactions.

\vspace{-4pt}
\begin{table}[h!]
\centering
\small
\caption{Baseline performance comparison for emotion and valence classification using different contextual inputs: current turn ($p_i^{(0)}$), previous interlocutor turn ($p_i^{(1)}$), and customer's prior turn for same-turn ($p_i^{(2a)}$) vs. future-turn ($p_i^{(2b)}$) prediction across Dutch (NL) and English (EN) dialogues (mean ± std over 5 folds).}
\label{tab:exp_baselines}
\begin{tabular}{
  l
  l
  S[table-format=1.4(1)]
  S[table-format=1.4(1)]
  S[table-format=1.4(1)]
}
\toprule
Setup & Lang. & {Accuracy} & {F1-macro} & {F1-weighted} \\
\midrule
$p_{i}^{(0)}(y \mid \Omega = \mathcal{E})$     & NL & 0.7438 \pm 0.0201 & 0.4256 \pm 0.0554 & 0.7042 \pm 0.0279 \\
                      & EN & 0.7389 \pm 0.0160 & 0.4378 \pm 0.0337 & 0.7063 \pm 0.0310 \\
\addlinespace[2pt]
$p_{i}^{(1)}(y \mid \Omega = \mathcal{E})$  & NL & 0.5318 \pm 0.0093 & 0.1670 \pm 0.0223 & 0.4550 \pm 0.0324 \\
                      & EN & 0.5313 \pm 0.0082 & 0.1669 \pm 0.0154 & 0.4525 \pm 0.0147 \\
\addlinespace[2pt]
$p_{i}^{(2a)}(y \mid \Omega = \mathcal{E})$     & NL & 0.7175 \pm 0.0091 & 0.4261 \pm 0.0450 & 0.6856 \pm 0.0267 \\
                     & EN & 0.7140 \pm 0.0109 & 0.4536 \pm 0.0204 & 0.6926 \pm 0.0205 \\
\addlinespace[2pt]
$p_{i}^{(2b)}(y \mid \Omega = \mathcal{E})$       & NL & 0.4852 \pm 0.0091 & 0.1310 \pm 0.0234 & 0.3926 \pm 0.0186 \\
                     & EN & 0.4912 \pm 0.0172 & 0.1198 \pm 0.0178 & 0.3845 \pm 0.0145 \\

\cmidrule(r){1-1}\cmidrule(r){2-2}\cmidrule(r){3-3}\cmidrule(r){4-4}\cmidrule{5-5}
$p_{i}^{(0)}(y \mid \Omega = \mathcal{V})$      & NL & 0.7620 \pm 0.0091 & 0.4308 \pm 0.0276 & 0.7433 \pm 0.0090 \\
                      & EN & 0.7614 \pm 0.0073 & 0.4536 \pm 0.0499 & 0.7457 \pm 0.0151 \\
\addlinespace[2pt]
$p_{i}^{(1)}(y \mid \Omega = \mathcal{V})$ & NL & 0.6494 \pm 0.0109 & 0.2811 \pm 0.0312 & 0.5988 \pm 0.0339 \\
                      & EN & 0.6369 \pm 0.0210 & 0.2603 \pm 0.0328 & 0.5798 \pm 0.0221 \\
\addlinespace[2pt]    
$p_{i}^{(2a)}(y \mid \Omega = \mathcal{V})$     & NL & 0.8005 \pm 0.0124 & 0.4156 \pm 0.0115 & 0.7836 \pm 0.0134 \\
                                                                 & EN & 0.7950 \pm 0.0087 & 0.4076 \pm 0.0261 & 0.7760 \pm 0.0104 \\
\addlinespace[2pt]
$p_{i}^{(2b)}(y \mid \Omega = \mathcal{V})$       & NL & 0.6421 \pm 0.0244 & 0.2201 \pm 0.0438 & 0.5512 \pm 0.0288 \\
                     & EN & 0.6189 \pm 0.0269 & 0.2377 \pm 0.0268 & 0.5534 \pm 0.0265 \\
\bottomrule
\end{tabular}
\end{table}

\begingroup
\begin{table}[t]
\centering
\small
\caption{Performance comparison of weighted fusion approaches ($\hat{p}_{i}^{(w)}$) combining other-to-future prediction (Model 1) with either persistence-based ($k=a$) or future-projection ($k=b$) variants of self-contextual prediction (Model 2) for emotion and valence inference across Dutch (NL) and English (EN) dialogues (mean ± std over 5 folds). Bold values indicate better performance between the two variants ($k=a$ vs. $k=b$) for the same target label space ($\mathcal{E}$ vs. $\mathcal{V}$) and language.}
\label{tab:exp_fusion}
\begin{tabular}{
l
l
S[table-format=1.4(1)]
S[table-format=1.4(1)]
S[table-format=1.4(1)]
}
\toprule
Setup & Lang. & {Accuracy} & {F1-macro} & {F1-weighted} \\
\midrule
$\hat{p}_{i}^{(w, k=a)}(y \mid \Omega = \mathcal{E})$ & NL & 0.5361 \pm 0.0203 & \bfseries0.1691 \pm 0.0238 & {\num{0.4549 \pm 0.0323}\footnotemark[13]} \\
& EN & 0.5290 \pm 0.0157 & 0.1740 \pm 0.0121 & {\num{0.4501 \pm 0.0231}\footnotemark[13]} \\
\addlinespace[2pt]
$\hat{p}_{i}^{(w, k=b)}(y \mid \Omega = \mathcal{E})$ & NL & \bfseries0.5492 \pm 0.0159 & 0.1610 \pm 0.0224 & 0.4549 \pm 0.0340 \\
& EN & \bfseries0.5488 \pm 0.0158 & \bfseries0.1792 \pm 0.0176 & \bfseries0.4667 \pm 0.0287 \\

\cmidrule(r){1-1}\cmidrule(r){2-2}\cmidrule(r){3-3}\cmidrule(r){4-4}\cmidrule{5-5}

$\hat{p}_{i}^{(w, k=a)}(y \mid \Omega = \mathcal{V})$ & NL & 0.6519 \pm 0.0153 & \bfseries0.2453 \pm 0.0376 & \bfseries0.5761 \pm 0.0410 \\
& EN & 0.6489 \pm 0.0111 & 0.2230 \pm 0.0141 & 0.5609 \pm 0.0214 \\
\addlinespace[2pt]
$\hat{p}_{i}^{(w, k=b)}(y \mid \Omega = \mathcal{V})$ & NL & \bfseries0.6548 \pm 0.0136 & 0.2332 \pm 0.0408 & 0.5678 \pm 0.0404 \\
& EN & \bfseries0.6593 \pm 0.0161 & \bfseries0.2236 \pm 0.0221 & \bfseries0.5615 \pm 0.0257 \\
\bottomrule
\end{tabular}
\footnotetext[13]{No value is boldface because both variants have identical F1-weighted scores.}
\end{table}

\begin{table}[t]
\vspace{5pt}
\centering
\small
\caption{Ablation study evaluating weighted fusion performance gains for emotion and valence classification across Dutch (NL) and English (EN) dialogues. Performance differences ($\Delta ± \text{SE}(\Delta)$) compare optimized weighted fusion variants (persistence-based ($k=a$) and future-projection ($k=b$)) against baseline models: other-to-future ($p_i^{(1)}$), same-turn persistence ($p_i^{(2a)}$), and future-projection ($p_i^{(2b)}$). Positive values indicate fusion outperforms the individual baseline model.}
\label{tab:exp_fusion_w_ablation}
\begin{tabular}{l l S[table-format=1.4(1)] S[table-format=1.4(1)] S[table-format=1.4(1)]}
\toprule
Model comparison & Lang. & {Accuracy} & {F1-macro} & {F1-weighted} \\
\midrule
$\hat{p}_{i}^{(w, k=a)}(y \mid \Omega = \mathcal{E}) - p_{i}^{(1)}(y)$ & NL & \phantom{-}0.0071 \pm 0.0126 & -0.0007 \pm 0.0151 & -0.0006 \pm 0.0196 \\
$\hat{p}_{i}^{(w, k=a)}(y \mid \Omega = \mathcal{E}) - p_{i}^{(1)}(y)$ & EN & \phantom{-}0.0005 \pm 0.0117 & -0.0107 \pm 0.0060 & -0.0171 \pm 0.0139 \\
\addlinespace[2pt]
$\hat{p}_{i}^{(w, k=a)}(y \mid \Omega = \mathcal{E}) - p_{i}^{(2a)}(y)$ & NL & \phantom{-}0.1763 \pm 0.0124 & \phantom{-}0.0333 \pm 0.0115 & \phantom{-}0.1193 \pm 0.0162 \\
$\hat{p}_{i}^{(w, k=a)}(y \mid \Omega = \mathcal{E}) - p_{i}^{(2a)}(y)$ & EN & \phantom{-}0.1575 \pm 0.0082 & \phantom{-}0.0281 \pm 0.0069 & \phantom{-}0.1068 \pm 0.0118 \\
\addlinespace[6pt]
$\hat{p}_{i}^{(w, k=b)}(y \mid \Omega = \mathcal{E}) - p_{i}^{(1)}(y)$ & NL & \phantom{-}0.0202 \pm 0.0109 & -0.0088 \pm 0.0145 & -0.0006 \pm 0.0200 \\
$\hat{p}_{i}^{(w, k=b)}(y \mid \Omega = \mathcal{E}) - p_{i}^{(1)}(y)$ & EN & \phantom{-}0.0175 \pm 0.0103 & -0.0242 \pm 0.0095 & -0.0117 \pm 0.0163 \\
\addlinespace[2pt]
$\hat{p}_{i}^{(w, k=b)}(y \mid \Omega = \mathcal{E}) - p_{i}^{(2b)}(y)$ & NL & \phantom{-}0.0597 \pm 0.0136 & \phantom{-}0.0281 \pm 0.0167 & \phantom{-}0.0610 \pm 0.0210 \\
$\hat{p}_{i}^{(w, k=b)}(y \mid \Omega = \mathcal{E}) - p_{i}^{(2b)}(y)$ & EN & \phantom{-}0.0528 \pm 0.0092 & \phantom{-}0.0722 \pm 0.0152 & \phantom{-}0.0958 \pm 0.0185 \\
\cmidrule(r){1-1}\cmidrule(r){2-2}\cmidrule(r){3-3}\cmidrule(r){4-4}\cmidrule{5-5}
$\hat{p}_{i}^{(w, k=a)}(y \mid \Omega = \mathcal{V}) - p_{i}^{(1)}(y)$ & NL & \phantom{-}0.0101 \pm 0.0077 & -0.0063 \pm 0.0284 & \phantom{-}0.0012 \pm 0.0240 \\
$\hat{p}_{i}^{(w, k=a)}(y \mid \Omega = \mathcal{V}) - p_{i}^{(1)}(y)$ & EN & -0.0039 \pm 0.0071 & -0.0164 \pm 0.0190 & -0.0109 \pm 0.0163 \\
\addlinespace[2pt]
$\hat{p}_{i}^{(w, k=a)}(y \mid \Omega = \mathcal{V}) - p_{i}^{(2a)}(y)$ & NL & \phantom{-}0.0575 \pm 0.0082 & -0.0089 \pm 0.0190 & \phantom{-}0.0215 \pm 0.0202 \\
$\hat{p}_{i}^{(w, k=a)}(y \mid \Omega = \mathcal{V}) - p_{i}^{(2a)}(y)$ & EN & \phantom{-}0.0579 \pm 0.0157 & -0.0183 \pm 0.0133 & \phantom{-}0.0175 \pm 0.0136 \\

\addlinespace[6pt]
$\hat{p}_{i}^{(w, k=b)}(y \mid \Omega = \mathcal{V}) - p_{i}^{(1)}(y)$ & NL & \phantom{-}0.0130 \pm 0.0076 & -0.0184 \pm 0.0281 & -0.0071 \pm 0.0269 \\
$\hat{p}_{i}^{(w, k=b)}(y \mid \Omega = \mathcal{V}) - p_{i}^{(1)}(y)$ & EN & \phantom{-}0.0082 \pm 0.0089 & -0.0415 \pm 0.0145 & -0.0267 \pm 0.0152 \\
\addlinespace[2pt]
$\hat{p}_{i}^{(w, k=b)}(y \mid \Omega = \mathcal{V}) - p_{i}^{(2b)}(y)$ & NL & \phantom{-}0.0071 \pm 0.0101 & \phantom{-}0.0394 \pm 0.0258 & \phantom{-}0.0321 \pm 0.0207 \\
$\hat{p}_{i}^{(w, k=b)}(y \mid \Omega = \mathcal{V}) - p_{i}^{(2b)}(y)$ & EN & \phantom{-}0.0292 \pm 0.0171 & \phantom{-}0.0217 \pm 0.0195 & \phantom{-}0.0270 \pm 0.0212 \\

\bottomrule
\end{tabular}
\end{table}
\endgroup

Building on these baselines, we investigated weighted fusion methods in Tables~\ref{tab:exp_fusion} and~\ref{tab:exp_fusion_w_ablation}. Our experimental results in Table~\ref{tab:exp_fusion} reveal both the potential and limitations of combining contextual approaches for cross-turn emotion and valence prediction. While the fusion methods achieve moderate performance levels (53-55\% accuracy for emotions, 65-66\% for valence), they remain well below the same-turn detection baseline (74\% emotion accuracy, 76\% valence accuracy), underscoring the fundamental difficulty of predicting affective states from contextual information alone. As for the baselines, valence prediction achieves superior performance compared to emotion prediction across all settings, likely due to the number of classes and corresponding class imbalance issues as evidenced in the low F1-macro scores (0.16-0.18 for emotions vs. 0.22-0.25 for valence). Moreover, future-projection fusion ($k=b$) consistently outperforms persistence-based fusion ($k=a$) in terms of accuracy, particularly for emotions, suggesting that modeling emotional dynamics provides more predictive signal than assuming simple persistence. However, F1 scores reveal a more nuanced picture, with $k=a$ sometimes achieving superior F1-macro and F1-weighted performance despite lower accuracy. The notably larger standard deviations for F1 metrics indicate higher variability across folds, likely reflecting these measures' sensitivity to class distribution variations in the presence of class imbalance.

The ablation analysis in Table~\ref{tab:exp_fusion_w_ablation} exposes a critical limitation underlying these fusion approaches: the performance gains are predominantly driven by a single superior component rather than genuine complementarity between prediction strategies. The dramatic asymmetry in performance differences reveals that the weighted fusion models achieve minimal improvements over the other-to-future baseline ($p_i^{(1)}$), with gains often approaching zero or even turning negative for F1-macro and F1-weighted (e.g., -0.0242±0.0095 F1-macro for English emotion prediction with $k=b$ fusion). In stark contrast, our weighted fusion models demonstrate substantial improvements when compared to the persistence-based ($p_i^{(2a)}$) and future-projection ($p_i^{(2b)}$) baselines, with accuracy gains reaching 0.1763±0.0124 and F1-weighted improvements of up to 0.1193±0.0162 for Dutch emotion prediction. These results reveal a complex performance profile: while achieving modest accuracy improvements, particularly for the future-projection variant ($k=b$), the approaches simultaneously show decreases in F1-macro and F1-weighted scores compared to individual components. These mixed patterns are difficult to interpret definitively given the relatively large standard deviations reported in Table~\ref{tab:exp_fusion_w_ablation}, which introduce substantial uncertainty around practical significance. However, the consistent replication of this asymmetric pattern across both languages and target variables suggests that the fusion mechanism is essentially functioning as a sophisticated model selection procedure rather than achieving meaningful integration of complementary predictive signals. The performance trade-off, where accuracy gains come at the cost of balanced class prediction, indicates that optimized weighted fusion may be heavily weighting toward the superior other-to-future approach ($p_i^{(1)}$) rather than leveraging true complementarity between contextual sources.

\begin{table}[h]
\centering
\small
\caption{Zero-shot prediction of future emotional state (emotion label $\mathcal{E}$ and valence $\mathcal{V}$) from context using the customer's prior turn $t_{i-2}$ and previous operator turn $t_{i-1}$, evaluated on Dutch (NL) and English (EN) dialogues (mean ± std over 5 folds).}
\label{tab:exp_zero_shot}
\begin{tabular}{
  l
  l
  S[table-format=1.4(1)]
  S[table-format=1.4(1)]
  S[table-format=1.4(1)]
}
\toprule
Setup & Lang. & {Accuracy} & {F1-macro} & {F1-weighted} \\
\midrule
zero-shot $(y \mid \Omega = \mathcal{E})$     & NL & 0.1586 \pm 0.0121 & 0.1289 \pm 0.0074 & 0.1817 \pm 0.0163 \\
                      & EN & 0.2140 \pm 0.0164 & 0.1537 \pm 0.0082 & 0.2484 \pm 0.0190 \\
                      
\cmidrule(r){1-1}\cmidrule(r){2-2}\cmidrule(r){3-3}\cmidrule(r){4-4}\cmidrule{5-5}

zero-shot $(y \mid \Omega = \mathcal{V})$  & NL & 0.3562 \pm 0.0125 & 0.2175 \pm 0.0070 & 0.3590 \pm 0.0125 \\
                      & EN & 0.3633 \pm 0.0104 & 0.2303 \pm 0.0034 & 0.3669 \pm 0.0081 \\
                      
\bottomrule
\end{tabular}
\end{table}

As for zero-shot emotion inference, the results in Table~\ref{tab:exp_zero_shot} reveal a pronounced performance gap to all supervised systems, especially for multiclass emotion prediction with 0.21 accuracy and 0.15 macro-F1 for Dutch and 0.16 accuracy and 0.13 macro-F1 for English. For valence, zero-shot performance is markedly higher (the F1 macro scores of zero-shot and weighted fusion are more similar) yet still substantially below trained models in terms of accuracy, reaching 0.36 accuracy and 0.22 macro-F1 for Dutch and 0.36 accuracy with 0.23 macro-F1 for English. This pattern suggests that coarse affective polarity is more recoverable from context than fine-grained emotion categories under instruction-only prompting. Taken together, these findings reinforce three conclusions: (1) instruction-only zero-shot inference is insufficient for future-turn emotion recognition in imbalanced, multiclass settings, (2) valence remains comparatively tractable but still lags far behind supervised cross-turn predictors and their weighted fusions, and (3) cross-linguistic behavior is consistent; English and Dutch exhibit similar zero-shot patterns across metrics, with English marginally better, suggesting that the shortfall arises from task formulation rather than language-specific artifacts.

\begin{table}[h]
\centering
\small
\caption{Per-class F1 (mean ± std) for Dutch (NL) and English (EN) emotion labels across three setups: direct detection at time ${t_i}$, prediction from context using weighted fusion, and prediction from zero-shot LLM. ``S'' denotes the class support (percentage of samples per class) and is shared across methods. The weighted fusion results are produced with the future-projection variant ($k=b$) of weighted fusion ($p_i^{(2b)}$).}
\label{tab:exp_emo_perclass}
\resizebox{\textwidth}{!}{
\begin{tabular}{l
                S[table-format=1.4(4)]
                S[table-format=1.4(4)]
                S[table-format=1.4(4)]
                S[table-format=1.4(4)]
                S[table-format=1.4(4)]
                S[table-format=1.4(4)]
                S[table-format=4.0]}
\toprule
Label & \multicolumn{2}{c}{Direct detection on $t_i$} & \multicolumn{2}{c}{Prediction with weighted fusion} & \multicolumn{2}{c}{Prediction with zero-shot LLM} & \multicolumn{1}{c}{S (\%)} \\
\cmidrule(r){2-3}\cmidrule(r){4-5}\cmidrule(r){6-7}
 & {NL} & {EN} & {NL} & {EN} & {NL} & {EN} & \\
\midrule
Anger          & 0.3770 \pm 0.1474 & 0.4011 \pm 0.0377 & 0.1605 \pm 0.0813 & 0.1863 \pm 0.0642 & 0.2370 \pm 0.0356 & 0.2699 \pm 0.0440 & 6.1 \\
Annoyance      & 0.1069 \pm 0.0591 & 0.1470 \pm 0.0755 & 0.0171 \pm 0.0343 & 0.1117 \pm 0.0609 & 0.1582 \pm 0.0391 & 0.1723 \pm 0.0278 & 7.8 \\
Confusion      & 0.0899 \pm 0.0817 & 0.1523 \pm 0.1053 & 0.0000 \pm 0.0000 & 0.0000 \pm 0.0000 & 0.0409 \pm 0.0135 & 0.0285 \pm 0.0078 & 2.0 \\
Desire         & 0.6333 \pm 0.0305 & 0.5908 \pm 0.0318 & 0.0188 \pm 0.0376 & 0.0620 \pm 0.0305 & 0.0575 \pm 0.0204 & 0.0831 \pm 0.0303 & 7.9 \\
Disapp. & 0.3939 \pm 0.1280 & 0.3908 \pm 0.0751 & 0.0121 \pm 0.0242 & 0.0000 \pm 0.0000 & 0.0906 \pm 0.0134 & 0.1053 \pm 0.0475 & 2.3 \\
Fear           & 0.0000 \pm 0.0000 & 0.0000 \pm 0.0000 & 0.0000 \pm 0.0000 & 0.0000 \pm 0.0000 & 0.0525 \pm 0.0776 & 0.0857 \pm 0.0858 & 0.7 \\
Gratitude      & 0.8957 \pm 0.0338 & 0.9078 \pm 0.0111 & 0.5271 \pm 0.0554 & 0.5058 \pm 0.0314 & 0.3390 \pm 0.0244 & 0.3344 \pm 0.0367 & 17.3\\
Joy            & 0.4752 \pm 0.1752 & 0.4991 \pm 0.0595 & 0.0163 \pm 0.0327 & 0.0464 \pm 0.0244 & 0.0103 \pm 0.0128 & 0.0058 \pm 0.0116 & 5.7 \\
Neutral        & 0.8589 \pm 0.0106 & 0.8512 \pm 0.0184 & 0.6972 \pm 0.0167 & 0.7011 \pm 0.0145 & 0.1737 \pm 0.0175 & 0.2979 \pm 0.0192 & 50.1 \\
\bottomrule
\end{tabular}
}
\end{table}

\begin{table}[h]
\centering
\small
\caption{Per-class F1 (mean ± std) for Dutch (NL) and English (EN) valence on a 5-point scale (1 very negative, 5 very positive) across three setups: direct detection at time ${t_i}$, prediction from context using weighted fusion, and prediction from zero-shot LLM. ``S'' denotes the class support (percentage of samples per class) and is shared across methods. The weighted fusion results are produced with the future-projection variant ($k=b$) of weighted fusion ($p_i^{(2b)}$).}
\label{tab:exp_val_perclass}
\begin{tabular}{l@{\hspace{1.1em}}
                S[table-format=1.4(4)]
                S[table-format=1.4(4)]
                S[table-format=1.4(4)]
                S[table-format=1.4(4)]
                S[table-format=1.4(4)]
                S[table-format=1.4(4)]
                S[table-format=4.0]}
\toprule
Valence & \multicolumn{2}{c}{Direct detection on $t_i$} & \multicolumn{2}{c}{Prediction with weighted fusion} & \multicolumn{2}{c}{Prediction with zero-shot LLM} & \multicolumn{1}{c}{S (\%)} \\
\cmidrule(r){2-3}\cmidrule(r){4-5}\cmidrule(r){6-7}
& {NL} & {EN} & {NL} & {EN} & {NL} & {EN} & \\
\midrule
Score 1 & 0.0686 \pm 0.0958 & 0.1027 \pm 0.0891 & 0.0000 \pm 0.0000 & 0.0226 \pm 0.0302 & 0.0139 \pm 0.0171 & 0.0555\pm 0.0328 & 3.4 \\
Score 2 & 0.4873 \pm 0.0503 & 0.5090 \pm 0.0765 & 0.1623 \pm 0.1202 & 0.1482 \pm 0.0467 & 0.3473 \pm 0.0150 & 0.3559 \pm 0.0189 & 16.3 \\
Score 3 & 0.8566 \pm 0.0076 & 0.8542 \pm 0.0062 & 0.7867 \pm 0.0118 & 0.7906 \pm 0.0128 & 0.3890 \pm 0.0153 & 0.3948 \pm 0.0103 & 65.0 \\
Score 4 & 0.7283 \pm 0.0373 & 0.7165 \pm 0.0172 & 0.2170 \pm 0.1232 & 0.1562 \pm 0.0738 & 0.3373 \pm 0.0230 & 0.3454 \pm 0.0290 & 14.4 \\
Score 5 & 0.0133 \pm 0.0267 & 0.0857 \pm 0.1714 & 0.0000 \pm 0.0000 & 0.0000 \pm 0.0000 & 0.0000 \pm 0.0000 & 0.0000 \pm 0.0000 & 0.9 \\
\bottomrule
\end{tabular}
\end{table}

These gaps in performance raise questions about where the errors concentrate. A per-class breakdown in Tables~\ref{tab:exp_emo_perclass} and~\ref{tab:exp_val_perclass} shows that the errors are not uniformly distributed but instead scale with label frequency and the semantic granularity of the annotation taxonomy, clarifying why zero-shot lags most on multiclass emotions while appearing comparatively stronger on valence. Crucially, the results also show a distinctive zero-shot shift away from \textit{neutral}: in contrast to supervised models, the zero-shot LLM produces many more non-neutral predictions and avoids over-weighting \textit{neutral} despite its 50.1\% support.
Consequently, while direct detection on the current turn retains the highest overall F1 and weighted fusion best preserves high-support classes like \textit{neutral} (0.70 NL/EN) and \textit{gratitude} (0.51 NL/0.53 EN), zero-shot redistributes probabilities toward non-neutral, negatively connotated emotions, yielding comparatively strong per-class F1 for \textit{anger} (0.24 NL/0.27 EN), \textit{annoyance} (0.16 NL/0.17 EN), \textit{disappointment} (0.09 NL/0.11 EN), and \textit{fear} (0.05 NL/0.09 EN), outperforming weighted fusion across all four and even surpassing the direct detection baseline for \textit{annoyance} (0.11 NL/0.15 EN) and \textit{fear} (0 NL/EN). At the same time, zero-shot remains weak on fine-grained or low-support categories such as \textit{joy} (0.01 NL/EN), \textit{desire} (0.06 NL/0.08 EN) and \textit{confusion} (0.04 NL/0.03 EN).

\section{Conclusion} \label{sec6:conclus}
We reflect upon our findings in light of the outlined ROs in Section~\ref{sec6:discuss}. Section~\ref{sec6:future} concludes this paper by discussing its limitations and directions for future research.

\subsection{Discussion of Findings with Respect to the Research Objectives}\label{sec6:discuss}
Concerning the evaluation of the Wizard of Oz technique as a corpus design for emotion research \textbf{(RO1)}, we found that the WOZ technique with operator-steered valence trajectories is a viable data collection method for in-domain emotion research. It combines experimental control with ecological characteristics (tasks, goals, turn structure, language use) and, unlike proprietary interactions and public-facing social media chats, provides full interaction context (including pre-chat conditions), consistent access to conversation endings, and systematically varied response strategies linked to targeted valence outcomes. In contrast, enterprise data remain largely inaccessible and public threads on platforms such as Twitter are frequently redirected to offline channels, which limits longitudinal analysis of emotion shifts and the downstream effects of operator strategies. The WOZ design deployed here closes these gaps by collecting bilingual, scenario-grounded dialogues enriched with operator-strategy labels, self versus third-party emotion perspectives, and participant profiling.

Temporal analyses show consistent emotion shifts with successful conversation-level steering (most clearly for negative targets). The label distribution remains dominated by neutral content, reflecting an intrinsic class imbalance in customer interactions that should be preserved rather than “corrected” in corpus design or evaluation. Linking operator tactics to subsequent affect shows objective, problem-focused strategies (explanation, request information/action, help online/offline) precede neutrality or gratitude, whereas suboptimal moves (miscomprehension, non-collaborative/inappropriate responses, irony) amplify anger, annoyance, disappointment, desire, and confusion. Some affective strategies (cheerfulness, gratitude) foster positive reciprocity, whereas apology or generic empathy sometimes underperform, consistent with Little et al.~\cite{Little2013_more}’s finding that problem-focused strategies outperform emotion-focused ones in customer emotion management.

As for our second objective of quantifying human performance and variation in emotion annotation \textbf{(RO2)}, our results in Section~\ref{sec4:emotion} showed reliability is moderate for multilabel categories and valence, lower for arousal, and weakest for dominance, mirroring established difficulties in affect annotation and prior findings where valence typically yields higher agreement than arousal and dominance~\cite{Wood2018,Labat2024}. These patterns align with broader evidence that categorical emotions with clear lexical cues (for example, gratitude) achieve higher inter-annotator agreement, whereas subtler or less frequent emotions are harder to annotate consistently. Comparing self-reports with third-party labels indicates a systematic perspective gap: mutual alignment is relatively higher for neutral, gratitude, anger, annoyance, and joy, while emotions like admiration, fear, and relief are often self-reported without external recognition, with additional confusions among gratitude, joy, and relief. This divergence shows that observable cues alone cannot always reliably reveal internal emotional states, so disagreement should be reported and modeled as signal rather than discarded as noise when building and evaluating affective systems.

Finally, we created machine learning emotion detection and forward-looking inference models on \corpus~\textbf{(RO3)}. Across supervised and zero-shot paradigms, results show a clear hierarchy: same-turn detection is strongest, next-turn inference from context is markedly harder, and instruction-only zero-shot lags behind, especially on multiclass emotions. Same-turn detection sets an empirical ceiling (Dutch: 74.4\% emotion accuracy; 76.2\% valence), predicting from the operator’s preceding turn is much weaker (53.2\% emotion accuracy; 0.17 F1-macro), and directly projecting the customer’s previous affect performs worst (48.5\%); valence is consistently easier than categorical emotions, with negligible Dutch–English gaps. Weighted late fusion of other-to-future and self-context signals yields some minimal performance increases over the standalone baselines ($\approx$ 53-55\% for emotions; 65-66\% for valence). Ablations indicate that most gains stem from a single stronger branch rather than truly complementary cues. Moreover, while accuracy mostly favors future-projection fusion, F1-macro and F1-weighted fluctuate depending on the language and often deteriorate under class imbalance.

Our final observations concern zero-shot inference: despite low overall scores (0.16 NL/21.4 EN emotion accuracy; 0.36 NL/EN valence accuracy), zero-shot predictions appear to redistribute probability away from the dominant neutral class and yield comparatively stronger per-class F1 on negatively valenced categories (e.g. anger, annoyance, disappointment, and fear) than the supervised fusion baselines, which could suggest some sensitivity to implicit affect despite limited outward expression. This aligns with probabilistic accounts of emotion inference in theory-of-mind, where observers integrate context and latent mental states to infer how someone is likely to feel~\cite{Ong2019}. Those inferred feelings need not manifest as overt textual expression in the next turn, which helps explain why zero-shot may ``sense'' negative experience that remains unexpressed and thus penalized under our expression-focused ground truth. The observed self-report vs. third-party perspective gap in inter-annotator agreement likewise shows that observable cues often fail to reveal internal states.

\subsection{Limitations and Future Research}\label{sec6:future}
Four limitations stand out. First, \corpus~lacks proprietary, real-world service chats, leaving potential realism gaps despite the WOZ design. Second, next-turn inference from short local windows remains difficult, and the late-fusion gains suggest dominance of a single branch rather than complementary cues, leaving substantial room for improvement to models that integrate longer discourse context, role dynamics, and causal structure across turns. Third, human and machine evaluation rely on an expression-focused ground truth, even though self-reports reveal perspective gaps for several emotions. Fourth, instruction-only zero-shot underperforms on multiclass emotions, but shows sensitivity to subtle negative affect, hinting at a mismatch between unexpressed internal states and observable emotion expression in text.

For future work, we propose replacing short-window late fusion with long-context, role-aware models that retrieve salient episodes across the dialogue and attach causal, turn-level attributions to operator strategies. Moreover, supervision should be perspectivist: first-party self-reports can be explicitly paired with third-party judgments to model valid disagreement and prevent label collapse, an idea supported by recent surveys and evidence showing misalignment between third-party readings and authors’ private emotional states, even when third parties are LLMs~\cite{li-etal-2025-third}. Finally, the current machine learning approaches can be extended beyond text to cover multimodal emotion prediction by integrating, e.g., speech prosody, facial micro-expressions, and lightweight physiological signals, which can disambiguate the occasional lack of expressed emotions in text.

\begin{acks}
This research was supported by the Flemish Government through the Research Program Artificial Intelligence (174K02325) and by the Research Foundation Flanders (FWO-Vlaanderen) under grant number 1S96324N.
\end{acks}

\clearpage
\printbibliography

\clearpage

\appendix

\section{Examples From the Corpus}\label{app:examples}
\begin{figure}[h!]
\begin{center}
\includegraphics[width=0.85\columnwidth]{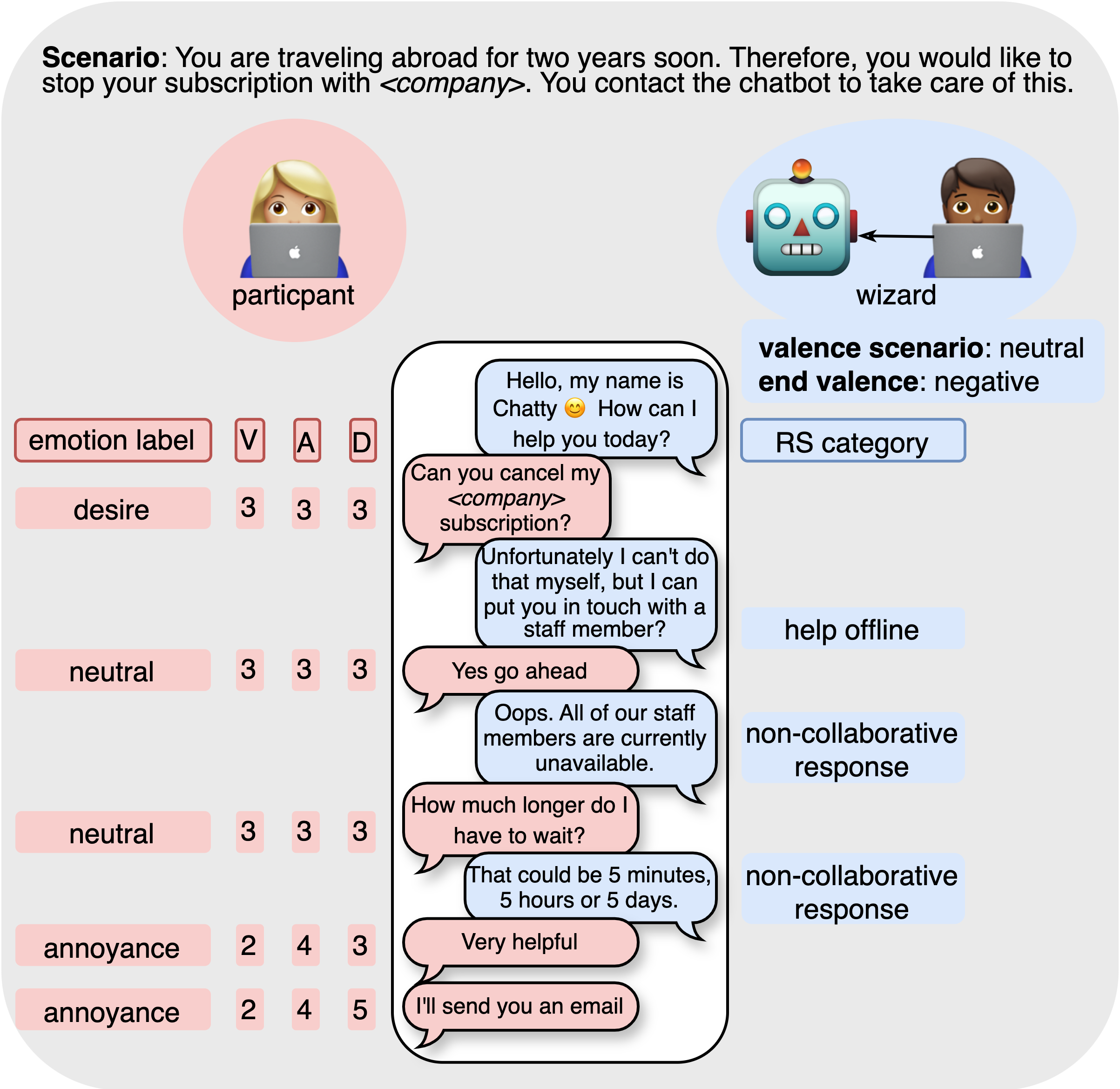}
\end{center}
\caption{Conversation in the \corpus~corpus. While the valence of the scenario is neutral, the wizard is instructed to navigate the interaction toward a negative end valence. Red boxes: participant messages are post-hoc labeled by annotators for emotion categories, valence (V), arousal (A), and dominance (D). Blue boxes: wizard replies are annotated in real-time with response strategies (RS category) by the wizards. The first wizard message is never annotated, as it is a scripted conversation starter that is shared across all interactions.}
\label{fig:ex_conv}
\Description{A single example dialogue is shown as a vertical sequence of colored message boxes alternating between participant and wizard. Participant utterances appear in red boxes with small tags indicating post-hoc emotion labels and numeric scores on a 5-point scale for valence (V), arousal (A), and dominance (D). Wizard utterances appear in blue boxes with labels indicating the response strategy (RS category) chosen in real time. The very first wizard turn is a fixed conversation starter and has no RS label. The scenario is neutral at the outset, but the wizard's instructions for this conversation specify steering toward a negative end valence.}
\end{figure}

This appendix showcases various aspects of \corpus~using actual data from the corpus. Table~\ref{tab:scenarios} gives four examples of scenarios in which conversations are grounded, one per economic sector. The table further displays the valence associated with each scenario. Figure~\ref{fig:ex_conv} contains a full conversation from our corpus. Participant messages of the conversation are annotated for emotion labels, valence, arousal, and dominance. Wizard replies are labeled with operator response strategies. More details on both annotation procedures are provided in Section~\ref{sec3:annot_rs}, Section~\ref{sec3:annot_emo}, and Appendix~\ref{app:annotation}.

\begin{table*}[h!]
\caption{Some scenarios of \corpus~in which conversations are grounded. Each scenario is linked to an economic sector and a valence (either negative or neutral).}
\label{tab:scenarios}
\centering
{\small
\begin{tabular}{p{2cm}p{10.5cm}p{1.5cm}}
\toprule
Economic \linebreak sector & Scenario & Valence\\
\midrule
Commercial aviation & You have just flown back from Miami, USA with \emph{$<$company$>$}. However, when you are waiting for your luggage, it turns out that it (unlike that of the other passengers) is not coming. Apparently, something went wrong and they lost your suitcase. Just when you had put a lot of souvenirs and personal items in your suitcase! Since you can't get anyone on the phone at customer service, you decide to contact the chatbot in the hope that it can help you further. & Negative\\
\cmidrule(r){1-1}\cmidrule(r){2-2}\cmidrule{3-3}
E-commerce & You recently ordered three pairs of pants and two shirts from \emph{$<$company$>$}. When your package arrives, you notice that one pair of pants is missing. You contact customer service to report this, but they don't believe you right away. However, you don't want to pay for something that never arrived. You therefore contact \emph{$<$company$>$}'s chatbot to explain your problem again. & Negative
\\
\cmidrule(r){1-1}\cmidrule(r){2-2}\cmidrule{3-3}
Online travel agencies & You have booked a stay via \emph{$<$company$>$} in Paris (France). A week before departure, it turns out that your travel companion has broken their foot. You contact the chatbot to ask if you can still cancel your stay. You would also like to know if this entails any extra costs. & Neutral
\\
\cmidrule(r){1-1}\cmidrule(r){2-2}\cmidrule{3-3}
Telecom & You are traveling abroad for two years soon. Therefore, you would like to stop your subscription with \emph{$<$company$>$}. You contact the chatbot to take care of this.
 & Neutral
\\
\bottomrule
\end{tabular}
}
\end{table*}

\clearpage
\section{Annotation Guidelines}\label{app:annotation}
Table~\ref{tab:emo_annot} holds an overview of the taxonomy used to label emotions in participant messages. Table~\ref{tab:rs_annot} contains the different response strategies that wizards used to classify their messages. Similar response strategies are clustered into techniques, a grouping used to expedite the annotation process. Each response strategy is accompanied by a definition and example.

\begin{table*}[h!]
\caption{Taxonomy to label emotions.}
\label{tab:emo_annot}
{\small
\begin{tabular}{p{3.7cm}p{10.7cm}}
\toprule
Emotion label & Description\\
\midrule
Anger & A strong feeling of displeasure, antagonism, and Disgust aroused by something unpleasant or offensive.\\
Annoyance & Mild frustration or irritation.\\
Disappointment & Sadness or displeasure caused by the nonfulfillment of one's hopes or expectations.\\
Fear & Afraid, worried, nervous, anxious, being concerned.\\
Confusion & Lack of understanding, uncertainty.\\
Desire & A strong feeling of wanting something or wishing something to happen.\\
Relief & Reassurance and relaxation following release from anxiety or distress.\\
Gratitude & A feeling of thankfulness and appreciation.\\
Joy & A feeling of contentment, happiness or pleasure.\\
Admiration & Finding something impressive, worthy of respect, or awe-inspiring.\\
Emotion label not found & The message conveys an emotion that does not fit any of the existing labels. Use this label only as a last resort, if there is no other option available.\\
Neutral & This message contains no emotion and is therefore objective.\\
\bottomrule
\end{tabular}
}
\end{table*}

\begin{table}[h!]
\caption{Response strategies and techniques for classifying wizard utterances, with definitions and examples.}
\label{tab:rs_annot}
\centering
{\footnotesize
\begin{tabular}{>{\raggedright\arraybackslash}p{0.13\linewidth}
                >{\raggedright\arraybackslash}p{0.20\linewidth}
                >{\raggedright\arraybackslash}p{0.30\linewidth}
                >{\raggedright\arraybackslash}p{0.22\linewidth}}
\toprule
\textbf{Technique} & \textbf{Response \linebreak strategy} & \textbf{Description} & \textbf{Example} \\
\midrule
\multirow[t]{4}{*}{Affective} & Apology & Denotes all types of apologies and regrets a wizard makes to customers. & I'm sorry for the inconvenience. \\
\cmidrule(r){2-2}\cmidrule(r){3-3}\cmidrule{4-4}
& Cheerfulness & Describes cheerful, happy, or funny responses (often with emojis). & You're welcome. Have a nice day! \\
\cmidrule(r){2-2}\cmidrule(r){3-3}\cmidrule{4-4}
& Empathy & Wizard empathizes by understanding, acknowledging, and mirroring emotions. & Oh, that's annoying! I understand your frustration. \\
\cmidrule(r){2-2}\cmidrule(r){3-3}\cmidrule{4-4}
& Gratitude & Wizard expresses gratitude to customers. Responses to customer gratitude fall under “cheerfulness.” & Thank you for coming to me for advice! \\
\midrule
\multirow[t]{5}{*}{Objective} & Explanation & Explains or motivates actions, events, products, services, or policies. & Your order should arrive later today. There was a problem with the supplier. \\
\cmidrule(r){2-2}\cmidrule(r){3-3}\cmidrule{4-4}
& Help offline & Refers customers to human operators or other departments. & This problem is too difficult for me. I will transfer you to an employee. \\
\cmidrule(r){2-2}\cmidrule(r){3-3}\cmidrule{4-4}
& Help online & Proposes to help customers. & I will solve your problem within 5 minutes. \\
\cmidrule(r){2-2}\cmidrule(r){3-3}\cmidrule{4-4}
& Request action & (1) Asks/advises customers to perform an action. \newline (2) Asks permission to undertake an action. & Have you turned your TV off and on again? \\
\cmidrule(r){2-2}\cmidrule(r){3-3}\cmidrule{4-4}
& Request \linebreak information & Asks for additional information to clarify or gather necessary details. & Could you rephrase the question? \\
\midrule
\multirow[t]{4}{*}{Suboptimal} & Inappropriate \linebreak response & Response that is not socially acceptable in the context. & Wow, sometimes I'm amazed at the stupid questions I get! \\
\cmidrule(r){2-2}\cmidrule(r){3-3}\cmidrule{4-4}
& Irony & Says something ironic given the context. & Pff, wat jammer zeg \#not… \\
\cmidrule(r){2-2}\cmidrule(r){3-3}\cmidrule{4-4}
& Miscomprehension & Fully or partially misunderstands the customer, leading to an off-target response. & \textit{[intent: cancel order]} Sure, I will extend your subscription for another year! \\
\cmidrule(r){2-2}\cmidrule(r){3-3}\cmidrule{4-4}
& Non-collaborative response & Does not cooperate or help the customer. & Customer service is not available today. \\
\midrule
Alternative & Other & Replies not covered by previous strategies. & / \\
\bottomrule
\end{tabular}
}
\end{table}

\clearpage
\section{Data Anonymization}\label{app:anonym}
To anonymize our data, we substituted all personally identifiable text and any company names with standardized placeholders. The following list details the categories of information that were replaced and their corresponding placeholders:

\begin{itemize}
    \item URL: $<$http\_url$>$
    \item Email address: $<$email$>$
    \item Street name (incl. number if present): $<$street$>$
    \item City: $<$city$>$
    \item Number (excl. dates): $<$number$>$
    \item First name: $<$firstname$>$
    \item Last name: $<$lastname$>$
    \item Payment details: $<$bank\_account\_number$>$
    \item Flight number: $<$flight\_number$>$
    \item Company names used in scenarios: $<$company$>$
    \item Competitors of $<$company$>$: $<$competitor$>$
    \item Postal service: $<$delivery\_service$>$
    \item All other companies that are not covered by the above categories: $<$other\_company$>$
\end{itemize}

\clearpage
\section{Corpus Analysis}\label{app:ar-dom_progression}

\begin{figure}[h!]
\begin{center}
\includegraphics[width=0.72\columnwidth]{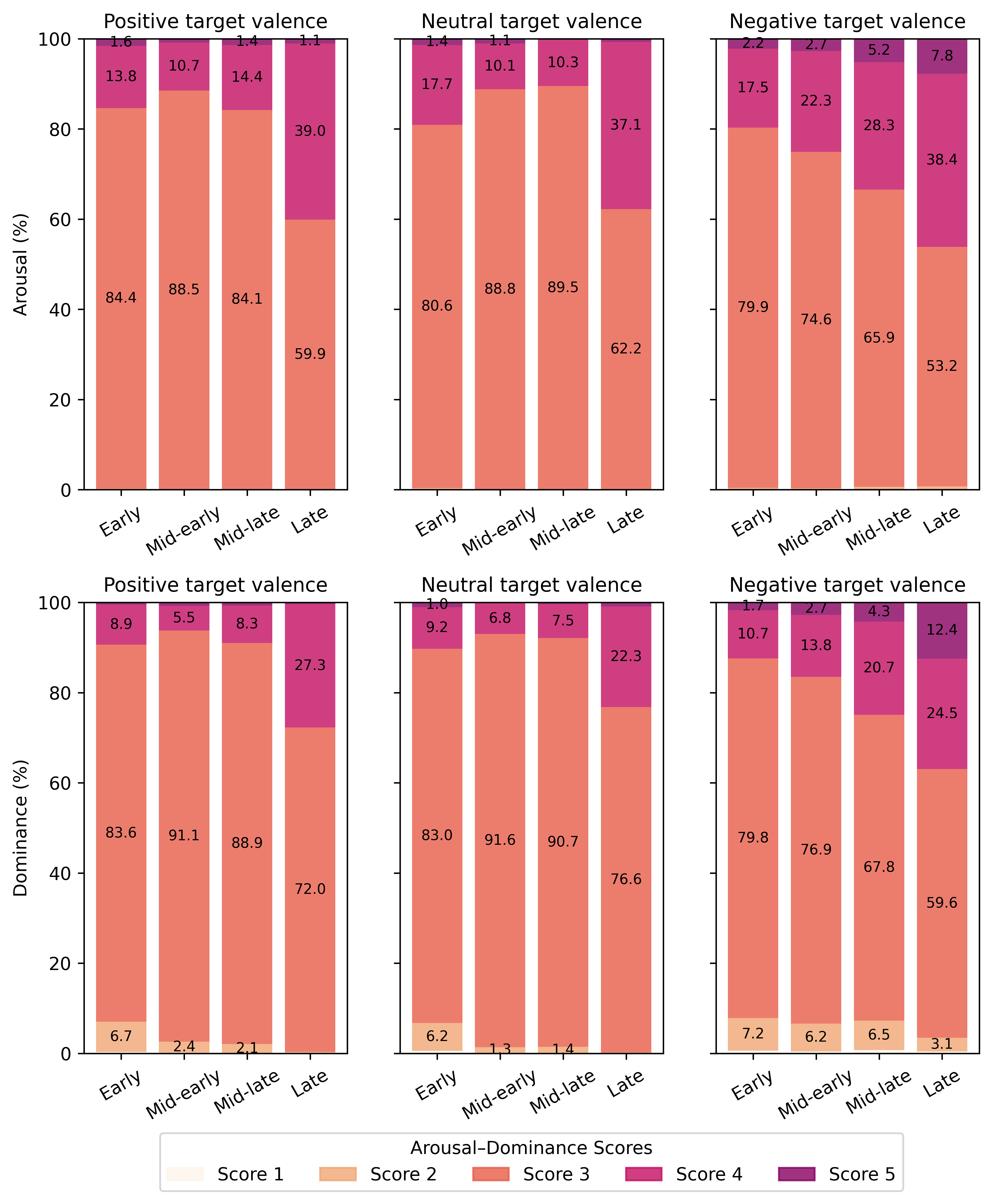}
\end{center}
\caption{Temporal progression of expressed arousal and dominance in interactions, grouped by target valence (positive, neutral, negative). Stacked bar charts show the distribution of scores from 1 (light shade) to 5 (dark shade) across four temporal bins. Percentage labels mark segments above 1\%. The four chronological bins depict arousal and dominance as they evolve over the course of the interactions.}
\label{fig:ar-dom_progress}
\Description{A set of stacked bar charts shows how arousal and dominance scores change over four chronological bins (early, mid-early, mid-late, late) for conversations assigned to positive, neutral, or negative target valence. Separate rows display arousal and dominance, each grouped by target condition (columns). Within each bar, five ordered segments represent score levels 1 to 5, rendered from light to dark shades; percentage labels appear on segments exceeding 1\%. Positive and neutral target conditions exhibit similar, relatively stable distributions across time, with moderate shifts in the ``late'' bin. In contrast, the negative target condition displays a noticeable drift toward higher arousal (more dark-shaded segments) as interactions progress, alongside a more polarized dominance pattern with increased proportions at both low and high ends, especially the high end. All panels share a consistent shading scale for scores, identical temporal binning, and aligned axes to allow side-by-side comparison of trajectories across targets and between arousal and dominance. The visualization emphasizes relative proportions within each time bin rather than raw counts, highlighting the emergence of heightened arousal and more extreme dominance under negative targets.}
\end{figure}

Figure~\ref{fig:ar-dom_progress} reports the temporal progression of arousal and dominance across interactions with distinct target valence values. While the trajectories for positive and neutral valence show minimal divergence, negative target valence consistently triggers heightened arousal and more extreme dominance scores at both ends of the spectrum.

\clearpage

\section{Details on Machine Learning Experiments}\label{app:exp}
This appendix reports per-fold data split sizes and the full bilingual prompt specifications used for zero-shot next-turn emotion inference.

\begin{table*}[h!]
\caption{Per-fold instance counts for the train, dev, and test splits (folds 1--5).}
\label{tab:data_splits_exp}
{\small
\begin{tabular}{lrrr}
\toprule
Fold & $|\text{Train}|$ & $|\text{Dev}|$ & $|\text{Test}|$\\
\midrule
1 & 4,944 & 944 & 1,461\\
2 & 4,929 & 1,064 & 1,356\\
3 & 4,868 & 1,012 & 1,469\\
4 & 4,861 & 1,075 & 1,413\\
5 & 4,627 & 1,262 & 1,460\\
\bottomrule
\end{tabular}
}
\end{table*}

\vspace{-4pt}
\begin{table*}[h]
\centering
\footnotesize
\caption{Structured overview of the next-turn emotion prediction prompts (Dutch vs. English) with categorical ($\mathcal{E}$) and valence ($\mathcal{V}$) setups that were used for zero-shot emotion inference.}
\label{tab:prompt-spec-en-nl}
\begin{tabularx}{\textwidth}{l X X}
\toprule
& NL & EN \\
\midrule
\multicolumn{3}{l}{\textit{Task framing}} \\
\cmidrule(r){1-1}\cmidrule(lr){2-3}
Role & Je bent een emotievoorspeller in een klantenservice-context. & You are an emotion predictor in a customer service setting. \\
\addlinespace[3pt]
Dialogue order & Gebruik de dialoogvolgorde:\newline klant t=i-2, operator t=i-1. & Use the dialogue order:\newline customer t=i-2, operator t=i-1. \\
\addlinespace[3pt]
Target turn & Voorspel de emotie die de klant zal uitdrukken in zijn volgende beurt (t=i). & Predict the emotion that the customer will express in their next turn (t=i). \\
\midrule
\multicolumn{3}{l}{\textit{Emotion labels ($\mathcal{E}$)}} \\
\cmidrule(r){1-1}\cmidrule(lr){2-3}
Label set & woede, ergernis, verwarring, verlangen, teleurstelling, angst, dankbaarheid, vreugde, neutraal & anger, annoyance, confusion, desire, disappointment, fear, gratitude, joy, neutral \\
\addlinespace[3pt]
System constraint & Geef exact één label uit: [label set].\newline Geen extra tekst. & Output exactly one label from: [label set].\newline No extra text. \\
\addlinespace[3pt]
User tail & Geef slechts één label terug uit de lijst. & Output only one label from the list. \\
\midrule
\multicolumn{3}{l}{\textit{Valence ($\mathcal{V}$)}} \\
\cmidrule(r){1-1}\cmidrule(lr){2-3}
Scale text & 1 = zeer negatief, 2 = negatief, 3 = neutraal, 4 = positief, 5 = zeer positief & 1 = very negative, 2 = negative, 3 = neutral, 4 = positive, 5 = very positive \\
\addlinespace[3pt]
System constraint & Schaal: [scale text]. Geef exact één geheel getal uit {1,2,3,4,5}. Geen uitleg, geen woorden, geen leestekens en geen spaties ervoor of erna. Geen kommagetallen of bereiken. & Scale: [scale text]. Return exactly one integer in {1,2,3,4,5}. No explanations, no words, no punctuation, and no whitespace before or after. Do not output floats or ranges. \\
\addlinespace[3pt]
User tail & Voorspel met de 5-punts numerieke schaal. [zelfde integer-only regel] & Predict using the 5-point numeric scale.\newline [same integer-only rule] \\
\midrule
\multicolumn{3}{l}{\textit{Context slots}} \\
\cmidrule(r){1-1}\cmidrule(lr){2-3}
Klant / Customer t=i-2 & Eerdere boodschap van de klant (t=i-2):\newline [self context text] & Customer's previous message (t=i-2):\newline [self context text]\\
\addlinespace[3pt]
Operator t=i-1 & Recente boodschap van de operator (t=i-1):\newline [opp context text] & Operator's recent message (t=i-1):\newline [opp context text] \\
\bottomrule
\end{tabularx}
\end{table*}
\clearpage

\end{document}